\documentclass{article}
    \PassOptionsToPackage{numbers, compress}{natbib}

    \usepackage[final]{neurips_2020}

\usepackage[utf8]{inputenc} 
\usepackage[T1]{fontenc}    
\usepackage{url}            
\usepackage{booktabs}       
\usepackage{amsfonts}       
\usepackage{nicefrac}       
\usepackage{microtype}      
\usepackage{mixedgcImp}     
\usepackage{adjustbox}
\usepackage{enumitem}

\newtheorem{definition}{Definition}
\newtheorem{theorem}{Theorem}
\newtheorem{lemma}{Lemma}

\newtheorem{corollary}{Corollary}

\title{Matrix Completion with Quantified Uncertainty \\ 
through Low Rank Gaussian Copula}

\author{%
  Yuxuan Zhao
    \\
  Cornell University\\
  \texttt{yz2295@cornell.edu} \\
  \And
  Madeleine Udell \\
  Cornell University\\
   \texttt{udell@cornell.edu} \\
}

\begin{document}

\maketitle

\begin{abstract}
Modern large scale datasets are often plagued with missing entries.
For tabular data with missing values,
a flurry of imputation algorithms solve for a complete matrix which minimizes some penalized reconstruction error.
However, almost none of them can estimate the uncertainty of its imputations.
This paper proposes a probabilistic and scalable framework for missing value imputation with 
quantified uncertainty.
Our model, the Low Rank Gaussian Copula, augments 
a standard probabilistic model, Probabilistic Principal Component Analysis, 
with marginal transformations for each column that allow the model to better match the distribution of the data.
It naturally handles Boolean, ordinal, and real-valued observations
and quantifies the uncertainty in each imputation.
The time required to fit the model scales linearly with the number of rows and the number of columns in the dataset.
Empirical results show 
the method yields state-of-the-art imputation accuracy across a wide range of data types, including those with high rank.
Our uncertainty measure predicts imputation error well: 
entries with lower uncertainty do have lower imputation error (on average).
Moreover, for real-valued data, the resulting confidence intervals are well-calibrated.
\end{abstract}

\section{Introduction}
Missing data imputation forms the first critical step of many data analysis pipelines;
indeed, in the context of recommender systems,
imputation itself is the task.
The remarkable progress in low rank matrix completion (LRMC)  \cite{candes2010matrix, keshavan2010matrix, recht2010guaranteed}
has led to wide use in collaborative filtering \cite{rennie2005fast},
transductive learning \cite{goldberg2010transduction}, 
automated machine learning \cite{yang2019oboe}, and beyond.
Nevertheless, 
reliable decision making requires one more step: 
assessing the uncertainty of the imputed entries.
While multiple imputation  \cite{rubin1996multiple,little2002statistical}  is a classical tool to quantify uncertainty,
its computation is often expensive and limits the use on large datasets.
For single imputation methods such as LRMC,
very little work has sought to quantify imputation uncertainty.
The major difficulty in quantifying uncertainty
lies in characterizing how the imputations depend on the observations
through the solution to a nonsmooth optimization problem.
\citet{chen2019inference} avoids this difficulty, providing confidence intervals for imputed real valued matrices,
by assuming isotropic Gaussian noise and a large signal-to-noise ratio (SNR).
However, these assumptions are hardly satisfied for most noisy real data.

The probabilistic principal component analysis (PPCA) model \cite{tipping1999probabilistic} provides a different approach to quantify uncertainty.
The PPCA model posits that the data in each row
is sampled iid from a Gaussian factor model.
In this framework, 
each missing entry has a closed form distribution conditional on the observations.
The conditional mean,
which is simply a linear transformation of the observations,
is used for imputation \cite{yu2009fast,ilin2010practical}.
However, the Gaussian assumption is unrealistic for most real datasets.

The Gaussian copula model presents a compelling alternative that 
enjoys the analytical benefits of Gaussians and yet fits real datasets well.
The Gaussian copula (or equivalently nonparanormal distribution) \cite{liu2009nonparanormal,fan2017high,feng2019high, hoff2007extending} can model real-valued, ordinal and Boolean data by transforming a latent Gaussian vector to match given marginal distributions.
Recently, \citet{zhao2020missing} proposed an imputation framework 
based on the Gaussian copula model and
empirically demonstrated state-of-the-art performance 
of Gaussian copula imputations on long skinny datasets.
However, 
their algorithm scales cubically in the number of columns,
which is too expensive for applications to large-scale datasets such as collaborative filtering and medical informatics.

\paragraph{Our contribution}
We propose a low rank Gaussian copula  model for imputation with quantified uncertainty.
The proposed model combines the advantages of PPCA and Gaussian copula:
the probabilistic description of missing entries allows for uncertainty quantification;
the low rank structure 
allows for efficient estimation from large-scale data;
and the copula framework provides the generality to
accurately fit real-world data.
The imputation proceeds in two steps:
first we fit the LRGC model,
and then we compute the distribution of the missing values 
separately for each row, conditional on the observed values in that row.
We impute the missing values with the conditional mean
and quantify their uncertainty with the conditional variance.
Our contributions are as follows.
\begin{enumerate}[leftmargin=*]
	\setlength{\parskip}{-1pt}
	\item We propose a probabilistic imputation method based on the low rank Gaussian copula model to impute real-valued, ordinal and Boolean data.
	The rank of the model is the only tuning parameter.
	\item We propose an algorithm to fit the proposed model 
	that scales linearly in the number of rows and the number of columns.
	Empirical results show our imputations provide state-of-the-art accuracy 
	across a wide range of data types, including those with high rank.
	\item We characterize how the mean squared error (MSE) of our imputations
	depends on the SNR.
	In particular, we show the MSE converges exponentially to the noise level in the limit of high SNR.
    \item We quantify the uncertainty of our estimates.
	Concretely, 
	we construct confidence intervals for imputed real values and 
	provide lower bounds on the probability of correct prediction for 
	imputed ordinal values.
	Empirical results show our confidence intervals are well-calibrated and our uncertainty measure predicts imputation error well:
	entries with lower estimated uncertainty do have lower imputation error (on average).
\end{enumerate}

\paragraph{Related work}
Although our proposed model has a low rank structure, 
it greatly differs from LRMC in that
the observations are assumed to be generated from,
but not equal to, a real-valued low rank matrix.
Many authors have considered generalizations of LRMC beyond real-valued low rank observations: 
to Boolean data \cite{davenport20141}, ordinal data \cite{lan2014matrix, bhaskar2016probabilistic, anderson2018xpca}, 
mixed data \cite{udell2016generalized, robin2020main},
data from an exponential family distribution \cite{gunasekar2014exponential}, 
and high rank matrices \cite{ganti2015matrix, ongie2017algebraic, fan2019online, fan2020polynomial}.
However, none of these methods can quantify the uncertainty of the resulting imputations.

Multiple imputation (MI)
requires repeating an imputation procedure many times 
to assess empirical uncertainty, 
often through bootstrap sampling \cite{josse2011multiple, audigier2017mimca} or Bayesian posterior sampling \cite{buuren2010mice} including probabilistic matrix factorization \cite{mnih2008probabilistic, salakhutdinov2008bayesian}.
The repeating procedure often leads to very expensive computation, especially for large datasets.
While variational inference can accelerate the process in some cases \cite{lim2007variational}, it may produce inaccurate results due to using overly simple approximation.
Moreover, \emph{proper} multiple imputation 
generally relies on strong distributional assumptions \cite{mnih2008probabilistic, salakhutdinov2008bayesian}. 
In contrast, our quantified uncertainty estimates are useful for a much broader family of distributions and can be computed as fast as a single imputation.
In addition, few MI papers explicitly explore the issue of \emph{calibration}: 
does MI sample variance predict imputation
accuracy? 
We find that the answer is usually no.
In contrast, our uncertainty metric is clearly correlated with imputation accuracy.

Some interesting new approaches \cite{carpentier2016constructing, carpentier2018adaptive} discuss constructing \emph{honest} confidence regions, which depends on some (possibly huge) hidden constants. 
However, these unknown hidden constants prevent its use in practice.
In contrast, our constructed confidence intervals are explicit.

Researchers from a Bayesian tradition have also studied the LRGC model with missing data \cite{murray2013bayesian, cui2019novel}.
However, the associated MCMC algorithms are expensive and do not scale to large-scale data. 


\section{Notation and background}
\label{gen_inst}
\paragraph{Notation}	Let $[p]=\{1,\ldots,p\}$ for $p\in \mathbb{N}^+$.
For a vector $\bx\in \mathbb{R}^{p}$ 
and a matrix $\bW\in \Rbb^{p\times k} (p>k)$,
with a set $I\subset [p]$,
we denote the subvector of $\bx$ with entries in $I$ as $\bx_{I}$,
and  the submatrix of $\bW$ with rows in $I$ as $\bW_I$.
Let $\bX\in \mathbb{R}^{n\times p}$ be a matrix 
 whose rows correspond to observations
and columns to variables.
We refer to the $i$-th row, $j$-th column, and $(i,j)$-th entry as $\bx^i, \bX_j$ and $\xij$, respectively.
Denote the vector $\ell_2$ norm as $||\cdot||_2$ and the matrix Frobenius norm as $||\cdot||_F$.

\paragraph{Gaussian copula}
The Gaussian copula models any random vector supported on an (entrywise) ordered set \cite{liu2009nonparanormal, hoff2007extending, fan2017high, feng2019high, zhao2020missing} including continuous (real-valued), ordinal, and binary observations, by transforming a latent Gaussian vector.
We follow the definition introduced in \cite{zhao2020missing}.
\begin{definition}
	We say a random vector $\bx=(x_1,\ldots,x_p)\in \mathbb{R}^{p}$ follows
	the Gaussian copula $\bx\sim \textup{GC}(\mathbf{\Sigma},\bigG)$
	with parameters $\mathbf{\Sigma}$ and $\bigG$
	if there exists a correlation matrix $\mathbf{\Sigma}$
	and elementwise monotone function
	$\bigG: \reals^p \to \reals^p$
	such that $\bigG(\bz):= (g_1(z_1),\ldots,g_p(z_p))=\bx$ for $\bz \sim \mathcal{N}_p(\mathbf{0},\mathbf{\Sigma})$.
	\label{definition:Gaussian_copula}
\end{definition}
Without loss of generality, we only consider increasing $g_j$ in this paper.
By matching the marginals of $x_j$ and $g_j(z_j)$, 
one can show $g_j(z)=F_j(\Phi^{-1}(z))$,
where $F_j$ is the cumulative distribution function (CDF) of $x_j$
for $j\in[p]$, and $\Phi$ is the standard Gaussian CDF.
Thus $x_j$ has continuous distribution if and only if $g_j$ is strictly monotone.
If $x_j$ is ordinal with $m$ levels, 
then $g_j$ is a step function with cut point set $\bS:1+\sum_{s\in\bS}\mathds{1}(z>s)$
\cite{zhao2020missing}. 
We focus on the case when $x_j$ for $j\in[p]$ are all continuous or all ordinal in this paper.
The extension to continuous and ordinal mixed data is natural as in \cite{zhao2020missing}.


The Gaussian copula model balances structure and flexibility by separately modeling the marginals $\bigG$ and the correlations $\mathbf{\Sigma}$ in the data:
assuming normality of the latent $\bz$ allows 
for constructing confidence intervals, while the 
flexibility of nonparametric $\bigG$ enables highly accurate fit to the data.
Define a set-valued inverse $g_j^{-1}(x_j):=\{z_j:g_j(z_j)=x_j\}$ 
and $\go^{-1}(\xobs):=\prod_{j\in \indexO}g_j^{-1}(x_j)$.
The inverse set $g_j^{-1}(x_j)$  is a point, an interval and $\Rbb$ for continuous,  ordinal and missing $x_j$, respectively.
The correlation matrix $\Sigma$ is estimated based on the observed information $\zobs\in\go^{-1}(\xobs)$ \cite{zhao2020missing}.

Interestingly, the Gaussian copula imputations do not significantly overfit even with $O(p^2)$ parameters in $\mathbf{\Sigma}$ \cite{zhao2020missing},
however, the fitting algorithm scales cubically in $p$ .
Without introducing low rank structures to $\mathbf{\Sigma}$, 
the Gaussian copula model is limited to skinny datasets with $p$ up to a couple of hundreds.

\section{Low rank Gaussian copula model}
\label{others}
We introduce our model, the model-based imputation and its quantified uncertainty in Section \ref{sec:model},
the estimation algorithm in Section \ref{sec:algorithm},
and analyze the MSE of our imputation estimator in Section \ref{sec:bound}.

\subsection{Model specification and the associated imputation}
\label{sec:model}
We propose a low rank Gaussian copula model that integrates 
the flexible marginals of the Gaussian copula model with 
the low rank structure of the PPCA model \cite{tipping1999probabilistic}.
To define the model, first consider a $p$-dimensional Gaussian vector $\bz \sim \textrm{PPCA}(\bW,\sigma^2)$
generated from the PPCA model:
\begin{equation}
	\bz = \bW\bt + \beps , \mbox{ where }\bt \sim \mathcal{N}_k(\bo,\Irm_k), \beps \sim \mathcal{N}_p(\bo,\sigma^2\Irm_p), \bt \mbox{ and }\beps \mbox{ are independent},
	 \label{model:ppca}
\end{equation}
where $\bW=[\bw_1,\ldots,\bw_p]^\intercal \in \Rbb^{p\times k}$ with $p>k$. 
We say $\bx$ follows the 
\emph{low rank Gaussian copula (LRGC) model}
if $\bx \sim \textup{GC}(\bW\bW^\intercal + \sigma^2 \Irm_p, \bigG)$ 
and $\bigG(\bz)=\bx$ for $\bz \sim \textrm{PPCA}(\bW,\sigma^2)$.

 To ensure that $\bx$ follows the Gaussian copula, 
 $\bz$ must have zero mean and unit variance in all dimensions. 
 Hence we require the covariance $\bW\bW^\intercal + \sigma^2\Irm_p$ to have unit diagonal: $||\bw_j||_2^2+\sigma^2=1$ for $j\in[p]$.
 We summarize the LRGC model in the following definition.
 
 \begin{definition}
 	We say a random vector $\bx\in \mathbb{R}^{p}$ follows
 	the low rank Gaussian copula $\bx\sim \lrgc$
 	with parameters $\bW\in \Rbb^{p\times k}(p>k), \sigma^2$ and $\bigG$
 	if (1) $\bigG$ is an elementwise monotonic function; (2) $\bW\bW^\intercal + \sigma^2 \Irm_p$ has unit diagonal; (3) $\bigG(\bz)=\bx$ for $\bz \sim \textrm{PPCA}(\bW,\sigma^2)$.
 	\label{definition: LRGC}
 \end{definition}

To see the generality of the LRGC model, 
suppose $\bX$ has iid rows $\bx^i\sim \lrgc$. Then
\begin{equation}
	\bX = \bigG(\bZ) = \bigG(\bT \bW^\intercal + \bE) := [g_1(\bZ_1),\ldots,g_p(\bZ_p)] =[g_1(\bT \bw_1+\bE_1),\ldots,g_p(\bT \bw_p + \bE_p)]
	\label{Eq:LRGC_factorization}
\end{equation}
where $\bZ, \bT,\bE$ have rows $\bz^i, \bt^i,\beps^i$, respectively, satisfying $\bz^i=\bW\bt^i + \beps^i$ and $\bigG(\bz^i)=\bx^i$ for $i\in [n]$.
While the latent normal matrix $\bZ$ has low rank plus noise structure,
the observation matrix $\bX$ can have high rank or ordinal entries 
with an appropriate choice of the marginals $\bigG$.
When all marginals of $\bigG$ are linear functions in $\Rbb$, 
the LRGC model reduces to the PPCA model.

Our method differs from LRMC and MI in that we treat one factor $\bW$ as model parameters, 
but the other factor $\bT$ as unseen random samples.
With estimated $\bW$, 
we analytically integrate over all $\bT$ to obtain the imputation and quantify uncertainty.
In contrast, LRMC and its generalization aim to estimate both factors $\bW$ and $\bT$ as model parameters,
which make it hard to quantify uncertainty;
MI treats both factors $\bW$ and $\bT$ as unseen random samples,
which make the computation, such as the posterior distribution, intractable and requires expensive sampling on large datasets.

\paragraph{Imputation}
Suppose we have observed a few entries $\xobs$ of a vector $\bx\sim \lrgc$
with known  $\bW$, $\sigma$ and $\bigG$.
We impute the missing entries $\xmis$ and 
quantify the imputation uncertainty.
We do not need to know the missing mechanism for imputation, 
so we defer a discussion to Section \ref{sec:algorithm}.

To impute $\xmis$, we need its conditional distribution given observation $\xobs$.
Since $\xmis = \bigG_{\indexM}(\zmis)$, 
we analyze the conditional distribution of $\zmis$, the latent Gaussian vector at $\indexM$.
From $\xobs$, we can be sure that the latent Gaussian vector 
$\zobs \in \go^{-1}(\xobs)$ lies in a known set based on the observed entries.
When all observed entries are continuous (i.e., $\go$ is strictly monotone),
the set $\go^{-1}(\xobs)$ is a singleton, so 
$\zobs$ is uniquely identifiable
and the conditional distribution $\zmis|\zobs$ is Gaussian.
However, for ordinal observations, 
$\go^{-1}(\xobs)$ is a Cartesian product of intervals.
The density of $\zmis|\xobs$ involves integrating $\zmis|\zobs$ over $\zobs\in \go^{-1}(\xobs)$,
which is intractable for $|\indexO| > 1$.
Fortunately, we can still estimate the mean and covariance of $\zmis|\xobs$, as stated in Lemma \ref{lemma: missing_moments},
upon which we construct imputation and quantify uncertainty.
All proofs appear in the supplement.
\begin{lemma}	\label{lemma: missing_moments}
	Suppose $\bx \sim \lrgc$ with observations $\xobs$ and missing entries $\xmis$.
	 Then for the latent normal vector $\bz$ satisfying $\bigG(\bz)=\bx$, with corresponding latent subvectors $\zobs$ and $\zmis$,
	\begin{align}
			\Erm[\zmis|\xobs] &= \Wmis \bM_{\indexO}^{-1}\Wobs^\intercal \Erm[\zobs|\xobs], \mbox{ where }\bM_{\indexO}=\sigma^2 \Irm_k + \Wobs^\intercal \Wobs \\
			 \Covrm[\zmis|\xobs]&=\sigma^2\Irm_{|\indexM|} +\sigma^2 \Wmis \bM_{\indexO}^{-1}\Wmis^\intercal + 		
			 \Wmis \bM_{\indexO}^{-1}\Wobs^\intercal\Covrm[\zobs|\xobs]\Wobs \bM_{\indexO}^{-1}\Wmis^\intercal
	\end{align}
\end{lemma}
For continuous $\xobs$,
the latent $\zobs$
is identifiable: $\Erm[\zobs|\xobs]=\go^{-1}(\xobs)$ and $\Covrm[\zobs|\xobs]=\bo$.
For ordinal $\xobs$, $\Erm[\zobs|\xobs]$ and $\Covrm[\zobs|\xobs]$ are the mean and covariance of a truncated normal vector,
determined by $\Wobs,\sigma^2$ and $\go^{-1}(\xobs)$.
We discuss the computation strategies in Section \ref{sec:algorithm}.

It is natural to impute $\xmis$ by mapping the conditional mean of $\zmis$ through the marginals $\gm$.
\begin{definition}[LRGC Imputation]
        Suppose $\bx \sim \lrgc$ with observations $\xobs$ and missing entries $\xmis$. We impute the missing entries as $\hat \bx_{\indexM} = \gm(\Erm[\zmis|\xobs])$ with $\Erm[\zmis|\xobs]$ in \cref{lemma: missing_moments}.
\label{definition:imputation}
\end{definition}

\paragraph{Imputation uncertainty quantification}
Can we quantify the uncertainty in these imputations?
Different from LRMC model which assumes a deterministic true value for missing locations, 
$\xmis$ (as well as $\zmis$) is random under the LRGC model.
Consequently, the error $\hat \bx_{\indexM} - \xmis$ is random and hence uncertain even with deterministic imputation $\hat \bx_{\indexM}$.
The uncertainty depends on the concentration of $\zmis$ around its mean $\Erm[\zmis|\xobs]$ and on the marginals $\gm$.
If $\gm$ is constant or nearly constant over the likely values of $\zmis$,
then with high probability the imputation is accurate.
Otherwise, 
the current observations cannot predict the missing entry well
and we should not trust the imputation.
Using this intuition, we may formally quantify the uncertainty in the imputations.
For continuous data, we construct confidence intervals using the normality of
$\zmis|\zobs=\go^{-1}(\xobs)$.
\begin{theorem}[Uncertainty quantification for continuous data]
	Suppose $\bx \sim \lrgc$ with observations $\xobs$ and missing entries $\xmis$
	and that $\bigG$ is elementwise strictly monotone.
	For missing entry $x_j$, 
	for any $\alpha\in (0,1)$,
	let $z^\star = \Phi^{-1}(1-\frac{\alpha}{2})$,
	the following holds with probability $1-\alpha$:
	\begin{equation}
x_j \in \left[g_j(\Erm[z_j|\xobs] - z^\star\Varrm[z_j|\xobs]),g_j(\Erm[z_j|\xobs] + z^\star\Varrm[z_j|\xobs])\right] =: [x_j^-(\alpha), x_j^+(\alpha)]
\label{eq:confidence_interval}
	\end{equation}
	where $\Erm[z_j|\xobs],\Varrm[z_j|\xobs]$ are given in \cref{lemma: missing_moments} with $\indexM$ replaced by $j$, for $j\in \indexM$.
	\label{theorem:confidence_interval}
\end{theorem}

 For ordinal data,
 we lower bound the probability of correct prediction $x_j = \hat x_j$ 
 using a sufficient condition that $z_j$ is sufficiently close to its mean $\Erm[z_j|\xobs]$.
General results in bounding $\Pr(|\hat x_j-x_j|\leq d)$ for any $d\in \mathbb{Z}$ appear in the supplement. 
Note a step function $g_j(z)$ with cut points set admits the form $\bS$: $g_j(z)=1+\sum_{s\in\bS}\mathds{1}(z>s)$.
\begin{theorem}[Uncertainty quantification for ordinal data]
	Suppose $\bx \sim \lrgc$ with observations $\xobs$ and missing entries $\xmis$
	and that the marginal $g_j$ is a step function with cut points $\bS_j$,
	for $j\in [p]$.
	For missing entry $x_j$
	and its imputation $\hat  x_j=g_j(\Erm[z_j|\xobs])$,
	\begin{equation}
	\Pr(\hat x_j=x_j)\geq 1 -{\Varrm[z_j|\xobs]}/{d^2_{j}}\quad \mbox{where }\quad d_{j}=\min\nolimits_{s\in \bS_j}\left|s-\Erm[z_j|\xobs]\right|,
	\label{eq:lower_bound_prob}
	\end{equation}
	and $\Erm[z_j|\xobs],\Varrm[z_j|\xobs]$ are given in \cref{lemma: missing_moments} with $\indexM$ replaced by $j$, for $j\in \indexM$.
	\label{theorem:uncertainty_ordinal}
\end{theorem}

To predict the imputation accuracy using the quantified uncertainty,
we develop a measure we call \emph{reliability}.
Entries with higher reliability are expected to have smaller imputation error.
We first motivate our definition of reliability.
For ordinal data, 
the reliability of an entry lower bounds the probability of correct prediction.
For continuous data,
our measure of reliability is designed so that reliable imputations have low normalized root mean squared error (NRMSE) under a certain confidence level $\alpha$.
NRMSE is defined as ${||P_{\Omega^c}(\bX-\hat\bX)||_F}/{||P_{\Omega^c}(\bX)||_F}$ 
for matrix $\bX$ observed on $\Omega$ and its imputed matrix $\hat \bX$.
Here $P_\Omega$ is projection onto the set $\Omega$: it sets entries not in $\Omega$ to 0.

Our definition of reliability uses Theorems \ref{theorem:confidence_interval}-\ref{theorem:uncertainty_ordinal} to ensure that reliable imputations have low error.
\begin{definition}[LRGC Imputation Reliability]
    	Suppose $\bX$ has iid rows $\xxi \sim \lrgc$ and is observed on $\Omega \subset [n] \times [p]$. 
    	Complete $\bX$ to $\hat \bX$ row-wise using \cref{definition:imputation}.
    	For each missing entry $(i,j) \in \Omega^c$,
    	define the \textit{reliability} of the imputation $\hat x^i_j$ as 
    	\begin{itemize}[leftmargin=*]
    	 	\setlength{\parskip}{0pt}
    	    \item (if $\bX$ is an ordinal matrix)
    	    the lower bound provided in Eq. (\ref{eq:lower_bound_prob});
    	    \item (if $\bX$ is a continuous matrix)
    	    ${||P_{\Omega^c\setminus (i,j)}(D_\alpha)||_F}/{||P_{\Omega^c\setminus(i,j)}(\hat \bX)||_F}$,
    	where the $(i',j')$-th entry of matrix $D_\alpha$ is the length of the confidence interval $\hat x^{i',+}_{j'}(\alpha) - \hat x^{i',-}_{j'}(\alpha)$ defined in Eq. (\ref{eq:confidence_interval}).
    	\end{itemize}
    	\label{definition:reliability}
\end{definition}
For continuous data,
the interpretation is that 
if the error after removing $(i_1,j_1)$ is larger than that after removing $(i_2,j_2)$,
then the imputation on $(i_1,j_1)$ is more reliable than that on $(i_2,j_2)$.
If continuous entries in different columns are measured on very different scales, 
one can also modify the definition to compute reliability column-wise.

Our experiments show this reliability measure positively correlates with imputation accuracy as measured by mean absolute error (MAE) for ordinal data and NRMSE for continuous data.
We also find for continuous data, the correlation is insensitive to $\alpha$ 
in a reasonable range; $\alpha=.05$ works well.

\subsection{Fitting the low rank Gaussian copula}
\label{sec:algorithm}
Suppose $\bX\in \Rbb^{n\times p}$ observed on $\Omega$ has iid rows $\xxi \sim \lrgc$ and $\bx^i$ has observations $\xio$ and missing entries $\xim$.
As in prior work
\cite{liu2009nonparanormal,guo2015graphical,zhao2020missing}, 
we estimate $\bigG$ by
matching it to the empirical distribution of observations.
To estimate $(\bW,\sigma^2)$,
we propose an EM algorithm that scales 
linearly in $n$ and $p$,
using ingredients from \cite{zhao2020missing, guo2015graphical} for the E-step,
and from \cite{ilin2010practical} for the M-step.
Due to space limit, 
we present essentials here and summarize in \cref{alg:LRGC}.
Details appear in the supplement.

\paragraph{Estimate the marginals}
\label{section:marginals_estimate}
We need the marginals $g_j$ for imputation and their inverse $g_j^{-1}$ to estimate the covariance.
Noticing $g_j$ maps standard normal data 
to $\bX_j$ which has CDF $F_j$,
thus we may estimate these functions by the empirical distribution of observations in $\bX_j$.
Concretely, recall that $g_j = F_j^{-1} \circ \Phi$ and $g_j^{-1} = \Phi^{-1} \circ F_j$.
We estimate $g_j^{-1}$ by replacing $F_j$ with the scaled empirical CDF 
$\frac{n}{n+1}\hat F_j$,
where the scaling $\frac{n}{n+1}$ is chosen to ensure finite output of $g_j^{-1}$.
Similarly, we estimate $g_j$ using the empirical quantiles of observations in $\bX_j$.
To ensure that the empirical CDF consistently estimates the true CDF, 
we assume the missing completely at random mechanism (MCAR).
It is possible to relax the it to missing at random (MAR) or even missing not at random by modeling either $F_j$ or the missing mechanism.
We leave this important work for the future.

\paragraph{EM algorithm for $\bW$ and $\sigma^2$}
Ideally, we would compute the maximum likelihood estimates (MLE) for the copula parameters $(\bW,\sigma^2)$ 
(under the likelihood in \cref{eq:observed_likelihood}),
which are consistent under the MAR mechanism
\cite[Chapter~6.2]{little2002statistical} as $n\xrightarrow[]{}\infty$.
However, the likelihood involves a Gaussian integral
that is hard to optimize.
Instead, we estimate the MLE
using an approximate EM algorithm.

The likelihood of $(\bW,\sigma^2)$  given observation $\xio$ is the integral
over the latent Gaussian vector $\zio$
that maps to $\xio$ under the marginal $\bigG_{\indexO_i}$.
Hence the observed log likelihood we seek to maximize is:
\begin{equation}
\ell_{\textup{obs}}(\bW,\sigma^2;\{\xio\}_{i=1}^n)	=\sum_{i=1}^n\log\int_{\zio \in \bigG_{\indexO_i}^{-1}(\xio)}\phi(\zio;\bo, \Wiobs \Wiobs^\intercal + \sigma^2\Irm_{|\indexO_i|})d\zio,
	\label{eq:observed_likelihood}
\end{equation}
where $\phi(\cdot;\bmu,\mathbf{\Sigma})$ denotes the Gaussian vector density with mean $\bmu$ and covariance $\mathbf{\Sigma}$.
Recall the decomposition $\bZ=\bT \bW^\intercal +\bE$ as in \cref{Eq:LRGC_factorization}.
If $\zio$ and $\bt^i$ are known,
the joint likelihood is simple:
\begin{equation}
    	\ell(\bW,\sigma^2;\{\xio,\zio,\bt^i\}_{i=1}^n)=\sum_{i=1}^n\log\left[ \phi(\zio;\Wiobs \bt^i,\sigma^2 \Irm_p)\, 
    	\phi(\bt^i;\bo,\Irm_k)\,
    	\mathds{1}_{\bigG_{\indexO_i}^{-1}(\xio)}(\zio) 
    	\right].
    	\label{Eq:joint_likelihood}
\end{equation}
Here define $\mathds{1}_A(x)=1$ when $x\in A$ and $0$ otherwise.
The maximizers $(\hat \bW, \hat \sigma)$ of \cref{Eq:joint_likelihood} are 
$\hat \bW=\argmin_{\bW}||P_{\Omega}(\bZ-\bT \bW^\intercal)||_F^2$ and 
$\hat \sigma^2 = ||P_{\Omega}(\bZ-\bT \hat \bW^\intercal)||^2_F/|\Omega|$.
Moreover, 
the problem is separable over the rows of $\hat \bW$:
to solve for the $j$-th row $\hat \bw_j^\intercal$, 
we use only $\zio,\bt^i$ for $i\in\Omega_j=\{i:(i,j)\in\Omega\}$.
Our EM algorithm treats the unknown $\zio,\bt^i$ as latent variables 
and $\xio$ as the observed variable.
Given an estimate $(\tilde \bW,\tilde \sigma^2)$,
the E-step computes the expectation  $\Ebb[||P_{\Omega}(\bZ-\bT \bW^\intercal)||_F^2]$ with respect to $\zio$ and $\bt^i$ conditional on $\xio$.
Throughout the section, we use $\Ebb$ to denote this conditional expectation.
The M-step is similar to when $\zio$ and $\bt^i$ are known.
\begin{description}[leftmargin=*]
	\setlength{\parskip}{0pt}
  \item[E-step] Calculate the expected likelihood $Q(\bW,\sigma^2;\tilde \bW, \tilde \sigma^2)=\mathbb{E}[\ell(\bW,\sigma^2;\{\xio,\zio,\bt^i\}_{i=1}^n)]$.
It suffices to compute $\Ebb[(\zio)^\intercal\zio], \Ebb[\bt^i(\zio)^\intercal]$ and $\Ebb[\bt^i(\bt^i)^\intercal ]$ as detailed below.
  \item[M-step] Let $\be_j\in\Rbb^p$ be the $j$th standard basis vector.
  The maximizers of $Q(\bW,\sigma^2;\tilde \bW, \tilde \sigma^2)$ are
 \begin{equation}
     \hat \bw_j^\intercal=\left(\be_j^\intercal\sum_{i\in \Omega_j}\Ebb[\zio(\bt^i)^\intercal] \right)\left(\sum_{i\in \Omega_j}\Ebb[\bt^i(\bt^i)^\intercal]\right)^{-1},
     \mbox{ }
          \hat{\sigma}^2=
\frac{\sum_{i=1}^n
\Ebb\left[||\zio-\hat \bW_{\indexO_i} \bt^i)||_2^2\right]}{\sum_{i=1}^n|\indiObs|}.
\label{Eq:M-step}
 \end{equation}
\end{description}
The maximizer $(\hat \bW, \hat \sigma^2)$ increase the observed likelihood in \cref{eq:observed_likelihood} compared to the initial estimate $(\tilde \bW, \tilde \sigma^2)$ \cite[Chapter~3]{mclachlan2007algorithm}.
To satisfy the unit diagonal constraints 
$|| \bw_j||_2^2 + \sigma^2=1$,
 we approximate the constrained maximizer by scaling the unconstrained maximizer shown in \cref{eq:projection} as in \cite{guo2015graphical,zhao2020missing}:
\begin{align}
\hat \sigma^2\leftarrow\hat \sigma^2_{\textup{new}} =\frac{1}{p}\sum_{j=1}^p\frac{\hat \sigma^2 }{||\hat \bw_j||_2^2 + \hat \sigma^2} ,\quad
\hat \bw_j\leftarrow\frac{\hat \bw_j}{||\hat \bw_j||_2}\cdot \sqrt{1-\hat\sigma^2_{\textup{new}}}.
\label{eq:projection}
\end{align}
We find this approximation works well in practice.

\paragraph{Computation}
We can express all expectations in the E-step using $\Erm[\zio|\xio]$ and $\Covrm[\zio|\xio]$:
	\begin{align}
\Ebb[\bt^i] &= \bM_{\indiObs}^{-1} \Wiobs^\intercal \Erm[\zio|\xio],
\quad\mbox{ where }\bM_{\indiObs}=\sigma^2 \Irm_k + \Wiobs^\intercal \Wiobs
\label{Estep:t-mean}. \\
\Ebb[\bt^i(\zio)^\intercal] &= \Ebb[\bt^i] \Erm[\zio|\xio]^\intercal +  \bM_{\indiObs}^{-1}\Wiobs^\intercal \Covrm[\zio|\xio]
\label{Estep:tz-mean}.  \\
\Ebb[\bt^i(\bt^i)^\intercal] &= \sigma^2 \bM_{\indiObs}^{-1}+\Ebb[\bt^i]	\Ebb[\bt^i]^\intercal+ \bM_{\indiObs}^{-1} \Wiobs^\intercal \Covrm[\zio|\xio] \Wiobs \bM_{\indiObs}^{-1}.
\label{Estep:tt-mean} 
\end{align}
For continuous data,
recall $\Erm[\zio|\xio]=\bigG_{\indiObs}^{-1}(\xio)$ and $\Covrm[\zio|\xio]=\bo$.
For ordinal data,
these quantities are the mean and covariance of a truncated normal vector, 
for each row $i$ separately at each EM iteration.
Direct computation \cite{bg2009moments} or sampling methods \cite{pakman2014exact} are expensive for large $n,p$.
We follow instead the fast iterative method in \cite{guo2015graphical,zhao2020missing}.
The intuition is that for each $j\in \indexO_i$, 
conditional on known $\bz_{\indexO_i\setminus\{j\}}$ and the constraint $\zij\in g_j^{-1}(\xij)$,
$\zij$ is univariate truncated normal with closed-form mean and variance.
Thus one can iteratively update the marginal mean and variance of $\zio|\xio$.
The rigorous formulation iteratively solves a nonlinear system satisfied by $\Erm[\zio|\xio]$ and by diagonals of $\Covrm[\zio|\xio]$ respectively, 
detailed in the supplement.
Estimating the off-diagonals of $\Covrm[\zio|\xio]$ efficiently for large $n,p$ is still an open problem: prior work approximates all off-diagonals as $0$ \cite{guo2015graphical,zhao2020missing}.
It is showed this diagonal approximation \cite{zhao2020missing}  yields more accurate and much faster parameters estimate than an alternative Bayesian algorithm without the approximation \cite{hoff2007extending}.
The approximation also reduces the computation for Eq. (\ref{Estep:tt-mean}) from $O(|\indObs|^2k)$ to $O(|\indObs|k^2)$. 

The computational complexity for each iteration is
$O(|\Omega| k^2 + nk^3 + pk^3)$, upper bounded by $O(npk^2)$. 
We find the method usually converges in fewer than $50$ iterations across our experiments.
See the Movielens 1M experiment in Section \ref{sec:experiments} for a run time comparison with state-of-the-are methods.

\renewcommand{\algorithmicrequire}{\textbf{Input:}}
\renewcommand{\algorithmicensure}{\textbf{Output:}}
\begin{algorithm}[h]
\caption{Imputation via low rank Gaussian copula fitting}
\label{alg:LRGC}
\begin{algorithmic}[1]
\Require
$\bX\in \Rbb^{n\times p}$ observed on $\Omega$, rank $k$, $t_{\text{max}}$.
\State Compute the empirical CDF $\hat F_j$ and empirical quantile function $\hat F_j^{-1}$ on observed $\bX_j$, for $j\in[p]$.
\State Estimate $\hat g_j = \Phi^{-1}\circ \frac{n}{n+1}\hat F_j$ and $\hat g_j^{-1} = \hat F_j^{-1}\circ \Phi$, for $j\in[p]$.
\State Initialize: $\bW^{(0)},(\sigma^2)^{(0)}$
\For{$t=1,2,\ldots, t_\textup{max}$}
\State E-step: compute the required conditional expectation using Eq. (\ref{Estep:t-mean}-\ref{Estep:tt-mean}).
\State M-step: update $\bW^{(t)},(\sigma^2)^{(t)}$ using Eq. (\ref{Eq:M-step}-\ref{eq:projection}).
\EndFor
\State Impute $\hat \bx^i_{\indexM_i}$ using Definition \ref{definition:imputation} for $i\in[n]$ with $\bigG=\hat \bigG, \bW=\bW^{(t_{\textup{max}})},\sigma^2=(\sigma^2)^{(t_{\textup{max}})}$.
\Ensure $\hat \bX$ with imputed $\hat x_j^i$ at $(i,j)\in \Omega^c$ and observed $\xij$ at $(i,j)\in \Omega$. 
\end{algorithmic}
\end{algorithm}

\subsection{Imputation error bound}
\label{sec:bound}
The imputation error consists of two parts: (1) the random variation of the error under the true LRGC model; (2) the estimation error of the LRGC model.
Analyzing the estimation error (2) is challenging for output from EM algorithm;
moreover, in our experiments we find that the imputation error 
can be attributed predominantly to (1), detailed in the supplement. Hence we leave (2) to future work.
To analyze the random variation of the error under the true LRCG model, 
we examine the MSE of $\hat \bx$ 
for a random row $\bx\sim\lrgc$ with fixed missing locations $\indexM$:
$\textup{MSE}(\hat \bx) = {||\bigG_{\indexM}(\hat \bz_{\indMis}) -  \bigG_{\indexM}(\bz_{\indMis})||_2^2}/{|\indMis|}$.
For continuous $\bx$ with strictly monotone $\bigG$, 
we must assume that $\bigG$ is Lipschitz to obtain a finite bound on the error.
With this assumption and assuming $\bW,\sigma^2$ fixed and known,
we can use the fact that $\zmis|\xobs$ is normal
to bound large deviations of the MSE.
\begin{theorem}
		Suppose subvector $\xobs$ of $\bx \sim \lrgc$ is observed and
		that all marginals $\bigG$ are strictly monotone with Lipschitz constant $L$.
		Denote the largest and the smallest singular values of $\bW'$ as $\lammax(\bW')$ and $\lammin(\bW')$.
		Then for any $t>0$, the imputed values $\hat \bx$ in \cref{definition:imputation} 
		satisfy
		\begin{equation*}
	\Pr\left[\textup{MSE}(\hat \bx) >
	L^2\sigma^2\left(\sqrt{1+\frac{1-\sigma^2}{\sigma^2+\lammin^2(\Wobs)} }+\sqrt{	2\left(1+\frac{\lammax^2(\Wmis)}{\sigma^2+\lammin^2(\Wobs)}\right)\frac{t}{|\indexM|}}\right)^2
	\right]\leq e^{-t}.
	\end{equation*}
	\label{theorem: bound_z_full}
\end{theorem}
Theorem \ref{theorem: bound_z_full} indicates the imputation error concentrates at
$\sigma^2+\frac{\sigma^2(1-\sigma^2)}{\sigma^2+\lammin^2(\Wobs)/\sigma^2}$ 
with an expansion multiplier $L^2$ due to the marginals $\bigG$.
The first term $\sigma^2$ represents the fraction of variance due to noise
and the second term is small when the SNR is large.
We also analyze the distribution of $\lammin^2(\Wobs)/\sigma^2$
under a random design to provide insight into when the error is small:
in \cref{corollary:subgaussian}, 
the second term vanishes with increasing observed length $|\indexO|$.
\begin{corollary}\label{corollary:subgaussian}
Under the conditions of Theorem \ref{theorem: bound_z_full}, 
further assume $\bW$ has independent sub-Gaussian rows $\bw_j$ with zero mean and covariance $\frac{1-\sigma^2}{k}\Irm_k$ for $j\in[p]$.
Suppose $c_1 k<{|\indexO|}< c_2|\indexM|$ for some constant $c_1>0$ depending on the sub-Gaussian norm of 
the scaled rows $\sqrt{\frac{k}{1-\sigma^2}}\bw_j$ 
and some absolute constant $c_2>0$. 
Then for some constant $c_3>0$ depending on $c_1,c_2$,
\begin{equation}
   \Pr\left[\textup{MSE}(\hat \bx) > L^2\sigma^2\left(1+K_{|\indexO|}\right)\right]\leq {c_3}/{|\indexO|},
   \quad \mbox{ where }K_{|\indexO|}=O\left(\sqrt{\log(|\indexO|)/|\indexO|}\right).
\end{equation}
\end{corollary}
See the supplement for definition of a sub-Gaussian vector.
Analyzing the imputation error for ordinal $\bx$ is much harder since $\zmis|\xobs$ is no longer Gaussian. 
We leave that for future work.


\section{Experiments and conclusion}
\label{sec:experiments}
Our experiments evaluate the imputation accuracy of \texttt{LRGC}, 
whether our reliability measure (denoted as \texttt{LRGC} reliability) can predict imputation accuracy well,
and the empirical coverage of our proposed confidence intervals.
For the second task, 
we evaluate the imputation on the $m\%$ entries with highest reliability for varying $m$.
We say a measure predicts imputation accuracy if
the imputation error on the $m\%$ entries is smaller for smaller $m$, 
i.e., it positively correlates with imputation accuracy.
We introduce below competitors for each task. 
Implementation details appear in the supplement.

For imputation comparison,
we implement LRMC methods \texttt{softImpute} \cite{mazumder2010spectral},
\texttt{GLRM} \cite{udell2016generalized} with $\ell_2, \ell_1$, bigger vs smaller (BvS, for ordinal data), hinge, and logistic loss.
We also implement the high rank matrix completion method \texttt{MMC} \cite{ganti2015matrix},
and \texttt{PPCA},
a special case of \texttt{LRGC} with Gaussian marginals.
To measure the imputation error, we use NRMSE for continuous data and 
MAE for ordinal data.

For reliability comparison, 
we compare with variance based reliability:
the imputation for a given missing entry is more reliable if it has smaller variance.
To obtain variance estiamte, 
we implement the PCA based MI method (denoted as \texttt{MI-PCA}) \cite{josse2011multiple, josse2016missmda}, 
and construct MI style uncertainty quantification for general imputation algorithms:
given an algorithm and incomplete $\bX$,
divide the observations into $N$ parts.
Then apply the algorithm $N$ times, 
each time additionally masking one part of the observations.
Compute the variance of the original missing entries across $N$ estimates.
We use $N=10$ in this paper.
We denote such methods as $\texttt{MI+Algorithm}$ for applied algorithm.

For confidence interval (CI) comparison, we compare with 
CI based on the \texttt{softImpute} imputation (denoted as \texttt{LRMC}) \cite{chen2019inference} ,
CI based on \texttt{PPCA}
and CI based on \texttt{MI-PCA}.
All constructed intervals except \texttt{LRGC} are derived assuming normality:
specifically, they all assume $\bX=\bX^\star + \bE$ for some low rank $\bX^\star$ and isotropic Gaussian error $\bE$.
In particular, their CIs are always symmetric around the imputed value,
while \texttt{LRGC} can yield asymmetrical CIs.
See the supplement for implementation details.

\paragraph{Synthetic experiments}
We consider three data types from LRGC:
continuous, 1-5 ordinal and binary.
We generate $\bW\in\Rbb^{p\times k}, \bT\in\Rbb^{n\times k},\bE\in \Rbb^{n\times p}$ with independent standard normal entries, then scale each row of $\bW$ such that $||\bw_j||_2^2+\sigma^2=1$.
Then generate $\bX=\bigG(\bZ)=\bigG(\bT \bW^\intercal + \sigma\bE)$ using $\bigG$ described below. 
Missing entries of $\bX$ are uniformly sampled. We set $n=500$ and $p=200$.
For continuous data, 
we use $g_j(z)=z$ to generate a low rank $\bX=\bZ$ and $g_j(z)=z^3$ to generate a high rank $\bX$.
We set $k=10, \sigma^2=0.1$ and the missing ratio as $40\%$.
For 1-5 ordinal data and binary data, 
we use step functions $g_j$ with random selected cut points.
We generate one $\bX$ with high SNR $\sigma^2=0.1$ and one $\bX$ with low SNR $\sigma^2=0.5$.
We set $k=5$ and the missing ratio as $60\%$.

We examine the sensitivity of each method to its key tuning parameter.
Both \texttt{LRGC} and \texttt{PPCA} do not overfit with large ranks.
We report results using the best tuning parameter in Table \ref{table:simulation_accuracy}.
The complete results and implementation details appear in the supplement.
All experiments are repeated $20$ times. 

\begin{table}
	\begin{center}
	\caption{Imputation error (NRMSE for continuous and MAE for ordinal) reported over 20 repetitions, with rank $r$ for available methods. \texttt{GLRM} methods are trained at rank $199$.}
	\label{table:simulation_accuracy}
	\resizebox{\columnwidth}{!}{
	\begin{tabular}{lccccc}
		\toprule
		Continuous  & LRGC & PPCA  & softImpute  & GLRM-$\ell_2$ & MMC\\
		\midrule
		Low Rank  & ${.347(.004)}, r=10$& $\mathbf{.338(.004)}, r=10$  & ${.371(.004)}, r=117$   &$.364(.003)$ & $.633(.007), r=130$\\
	High Rank     & $\mathbf{.517(.011)}, r=10$& $.690(.010), r=10$ & ${.703(.005)}, r=104$  & $.696(.006)$ & $.824(.011), r=137$\\
	\midrule
	1-5 ordinal  & LRGC & PPCA & softImpute & GLRM-BvS & GLRM-$\ell_1$\\
	\midrule
		High SNR   & $\mathbf{.358(.008)}, r=5$& $.501(.010), r=6$ & $.582(.011), r=83$ & $.407(.007)$ & $.689(.010)$\\
		Low SNR  & $\mathbf{.788(.013)}, r=5$ & $.863(.013), r=5$ & ${.951(.015)}, r=38$   &$.850(.011)$ &  $1.027(.020)$\\
		\midrule
		Binary	& LRGC & PPCA & softImpute & GLRM-hinge & GLRM-logistic \\
			\midrule
High SNR    & $\mathbf{.103(.003)}, r=5$ & $.116(.002), r=6$ &$.136(.003), r=71$ &$.140(.002)$  & $.117(.002)$ \\
Low SNR   & $\mathbf{.205(.006)}, r=5$ &  $.208(.005), r=5$ & $.234(.007, r=61$ &$.226(.006)$  & $.217(.005)$ \\
		\bottomrule
	\end{tabular}
	}
		\end{center}
\end{table}

Shown in Table \ref{table:simulation_accuracy},
\texttt{LRGC} performs the best in all but one settings.
The improvement is significant for high rank continuous data.
For low rank continuous data, \texttt{PPCA} performs the best as expected since the model is correctly specified.
The slightly larger error of \texttt{LRGC} is due to the error in estimating a nonparametric marginal $\bigG$.
Notice both \texttt{LRGC} and \texttt{PPCA}  admit much smaller rank as best parameter.

\begin{figure}
 	\centering
 	\includegraphics[width=\linewidth]{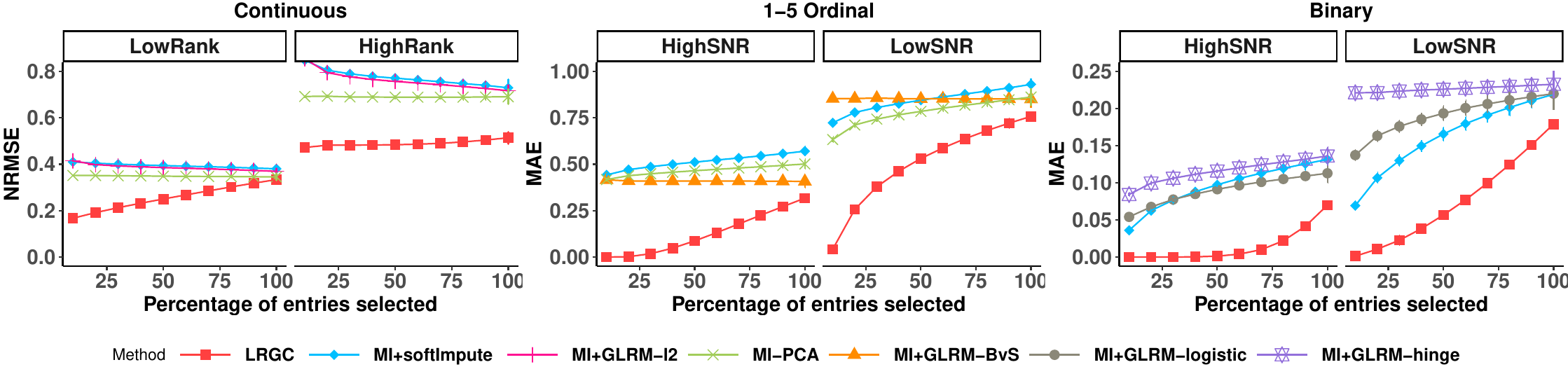}
 	\caption{Imputation error on the subset of $m\%$ entries for which method’s associated uncertainty metric indicates highest reliability, reported over $20$ repetitions (error bars almost invisible).}
 	\label{fig:uncertainty}
 \end{figure}

Shown in Figure \ref{fig:uncertainty},
 \texttt{LRGC} reliability predicts the imputation accuracy well:
entries with higher reliability (smaller $m$) have higher accuracy.
In contrast, 
entries with higher variance based reliability can have lower accuracy.
Even when the variance based reliability predicts accuracy,
\texttt{LRGC} reliability works better:
the error over selected entries using variance based reliability is much larger 
than that of \texttt{LRGC} reliability when a small percentage of entries $m$ are selected.
\texttt{LRGC} reliability can even find entries with error near $0$ 
from very noisy (low SNR 1-5 ordinal and binary) data.
\texttt{LRGC} reliability better predicts imputation error for easier imputation tasks (lower rank and higher SNR).
Predicting NRMSE is challenging,
since imputing continuous data is in general harder than imputing ordinal data.
In fact, we show in the supplement that as the number of levels of the ordinal variable increases,
the shape of the error vs reliability curve matches that of continuous data.

The results on confidence intervals appear in \cref{table:confidence_interval}.
Notice constructing \texttt{MI-PCA} intervals is much more expensive than all other methods.
For low rank Gaussian data, 
\texttt{PPCA} confidence intervals achieve the highest coverage rates with smallest length as expected,
since the model is correctly specified.
\texttt{LRGC} confidence intervals have slightly smaller coverage rates due to the error in estimating a nonparametric marginal $\bigG$.
For high rank data, 
the normality and the low rank assumption do not hold, 
so all other constructed confidence intervals but \texttt{LRGC} are no longer theoretically valid.
Notably, \texttt{LRGC} confidence intervals for more challenging high rank data achieves the same empirical coverage rates as that for low rank data.
The longer interval is due to the expanding marginal transformation $g_j(z)=z^3$.
While all other confidence intervals have visually good coverage rates, 
their interval lengths are much larger than \texttt{LRMC} confidence intervals,
which limits utility.

\begin{table}
	\centering
	\caption{$95\%$ Confidence intervals on synthetic continuous data over $20$ repetitions.}
	\label{table:confidence_interval}
\resizebox{0.8\columnwidth}{!}{
	\begin{tabular}{lcccc}
		\toprule
 Low Rank Data	& LRGC   & PPCA& LRMC  & MI-PCA \\
		\midrule
Empirical coverage rate	&  $0.927(.002)$   & $0.940(.001)$& $0.878(.006)$ & $0.933(.002)$ \\
Interval length & $1.273(.004)$  & $1.264(.004)$& $1.129(.015)$ & $1.267(.004)$\\
Run time (in seconds) & $6.9(.5)$  & $3.4(.7)$ & $2.7(.4)$& $189.8(15.4)$\\
\midrule
 High Rank Data 	& LRGC    & PPCA& LRMC & MI-PCA  \\
 \midrule
	Empirical coverage rate	& $0.927(.002)$   & $0.943(.002)$& $0.925(.004)$ & $0.948(.002)$ \\
	Interval length	& $3.614(.068)$ & $9.086(.248)$ & $6.546(.191)$ & $9.307(.249)$\\
	Run time (in seconds) & $7.2(1.2)$& $0.4(.1)$ 	& $3.1(.6)$  & $220.0(30.2)$ \\
		\bottomrule
	\end{tabular}
	}
\end{table}
 
\begin{figure}
 	\centering
 	\includegraphics[width=0.6\linewidth]{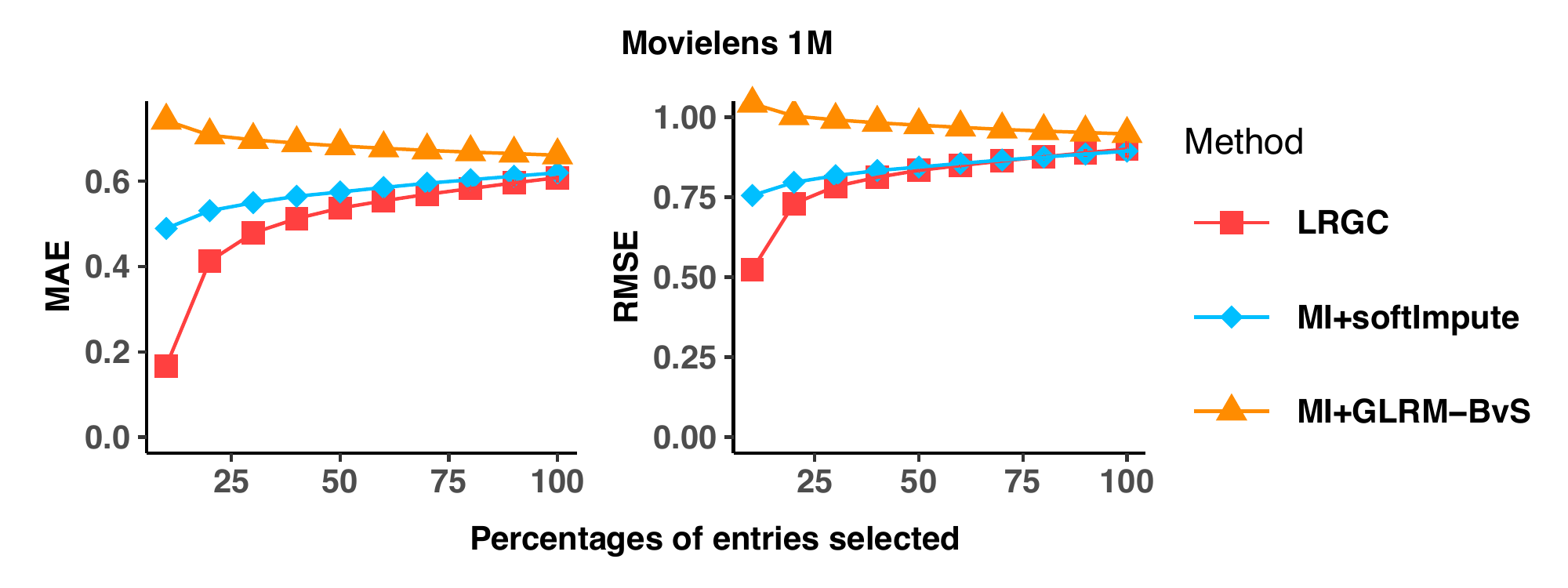}
 	\caption{Imputation error on the subset of $m\%$ entries for which method’s associated uncertainty metric indicates highest reliability, reported over $5$ repetitions (error bars almost invisible).}
 	\label{fig:uncertainty_real}
 \end{figure}

\paragraph{MovieLens 1M dataset}
We sample the subset of the MovieLens1M data \cite{harper2015movielens} consisting of $2514$ movies with at least $50$ ratings from $6040$ users.
We use $80\%$ of observation as training set, $10\%$ as validation set,
and $10\%$ as test set,
repeated $5$ times.
On a laptop with Intel-i5-3.1GHz Core
and 8 GB RAM,
\texttt{LRGC} (rank $10$) takes $38$ mins in \texttt{R}, 
\texttt{softImpute} (rank $201$) takes $93$ mins in \texttt{R}, 
and \texttt{GLRM-BvS} (rank $200$) takes $25$ mins in \texttt{julia}.
\texttt{MI-PCA} cannot finish even a single imputation in 3 hours in R.
We plot the imputation error versus reliability in Figure \ref{fig:uncertainty_real}.
The value at $m=100$ is the overall imputation error.
All methods have similar overall error.
The variance based reliability with \texttt{GLRM-BvS} cannot predict imputation accuracy.
In practice, collaborative filtering methods usually recommends very few entries to users.
In this setting, \texttt{LRGC} reliability predicts 
imputation accuracy much better than variance based reliability with \texttt{softImpute}.

\paragraph{Conclusion}
This paper develops a low rank Gaussian copula for matrix completion and 
quantifies the uncertainty of the resulting imputations. 
Numerical results show the superiority of our imputation algorithm and the success of our uncertainty measure for predicting imputation accuracy.
Quantifying imputation uncertainty can improve algorithms for recommender systems,
provide accurate confidence intervals for analyses of scientific and survey data,
and enable new active learning strategies.

\section*{Broader Impact}
In principle, 
this paper may benefit any research or practical projects that  requires imputing missing values, especially on large scale datasets.
The quantified uncertainty may serve as a step to exclude unreliable imputed entries or provide confidence levels after imputation.
In areas in which unreliable imputation could have adverse effect on people's lives, such as healthcare datasets, quantified uncertainty is important to aid related decision making.
On the other hand, if model assumptions are not met and the resulting quantified uncertainty estimate is not accurate,
it may incorrectly lead a practitioner to trust certain entries and distrust others entries when the opposite is true.
This problem may be mitigated by running some experiments on a validation set to check whether the proposed method works well on the particular dataset.

\section*{Acknowledgement}
We gratefully acknowledge support from
NSF Awards IIS-1943131 
and CCF-1740822, 
the ONR Young Investigator Program, %
DARPA Award FA8750-17-2-0101, 
the Simons Institute, 
Canadian Institutes of Health Research, 
and Capital One.
We thank Xin Bing,
David Bindel and Thorsten Joachims for helpful discussions.
Special thanks go to Xiaoyi Zhu for help in producing our figures.

\bibliographystyle{plainnat}
\bibliography{neurips_2020}

\begin{thebibliography}{50}
\providecommand{\natexlab}[1]{#1}
\providecommand{\url}[1]{\texttt{#1}}
\expandafter\ifx\csname urlstyle\endcsname\relax
  \providecommand{\doi}[1]{doi: #1}\else
  \providecommand{\doi}{doi: \begingroup \urlstyle{rm}\Url}\fi

\bibitem[Anderson-Bergman et~al.(2018)Anderson-Bergman, Kolda, and
  Kincher-Winoto]{anderson2018xpca}
Clifford Anderson-Bergman, Tamara~G Kolda, and Kina Kincher-Winoto.
\newblock {XPCA}: Extending {PCA} for a combination of discrete and continuous
  variables.
\newblock \emph{arXiv preprint arXiv:1808.07510}, 2018.

\bibitem[Audigier et~al.(2017)Audigier, Husson, and Josse]{audigier2017mimca}
Vincent Audigier, Fran{\c{c}}ois Husson, and Julie Josse.
\newblock Mimca: multiple imputation for categorical variables with multiple
  correspondence analysis.
\newblock \emph{Statistics and computing}, 27\penalty0 (2):\penalty0 501--518,
  2017.

\bibitem[BG and Wilhelm(2009)]{bg2009moments}
Manjunath BG and Stefan Wilhelm.
\newblock Moments calculation for the double truncated multivariate normal
  density.
\newblock \emph{Available at SSRN 1472153}, 2009.

\bibitem[Bhaskar(2016)]{bhaskar2016probabilistic}
Sonia~A Bhaskar.
\newblock Probabilistic low-rank matrix completion from quantized measurements.
\newblock \emph{The Journal of Machine Learning Research}, 17\penalty0
  (1):\penalty0 2131--2164, 2016.

\bibitem[Bing et~al.(2020)Bing, Ning, and Xu]{bing2020adaptive}
Xin Bing, Yang Ning, and Yaosheng Xu.
\newblock Adaptive estimation of multivariate regression with hidden variables.
\newblock \emph{arXiv preprint arXiv:2003.13844}, 2020.

\bibitem[Buuren and Groothuis-Oudshoorn(2010)]{buuren2010mice}
S~van Buuren and Karin Groothuis-Oudshoorn.
\newblock mice: Multivariate imputation by chained equations in r.
\newblock \emph{Journal of statistical software}, pages 1--68, 2010.

\bibitem[Candes and Plan(2010)]{candes2010matrix}
Emmanuel~J Candes and Yaniv Plan.
\newblock Matrix completion with noise.
\newblock \emph{Proceedings of the IEEE}, 98\penalty0 (6):\penalty0 925--936,
  2010.

\bibitem[Carpentier et~al.(2016)Carpentier, Klopp, and
  L{\"o}ffler]{carpentier2016constructing}
Alexandra Carpentier, Olga Klopp, and Matthias L{\"o}ffler.
\newblock Constructing confidence sets for the matrix completion problem.
\newblock In \emph{Conference of the International Society for Non-Parametric
  Statistics}, pages 103--118. Springer, 2016.

\bibitem[Carpentier et~al.(2018)Carpentier, Klopp, L{\"o}ffler, Nickl,
  et~al.]{carpentier2018adaptive}
Alexandra Carpentier, Olga Klopp, Matthias L{\"o}ffler, Richard Nickl, et~al.
\newblock Adaptive confidence sets for matrix completion.
\newblock \emph{Bernoulli}, 24\penalty0 (4A):\penalty0 2429--2460, 2018.

\bibitem[Chen et~al.(2019)Chen, Fan, Ma, and Yan]{chen2019inference}
Yuxin Chen, Jianqing Fan, Cong Ma, and Yuling Yan.
\newblock Inference and uncertainty quantification for noisy matrix completion.
\newblock \emph{Proceedings of the National Academy of Sciences}, 116\penalty0
  (46):\penalty0 22931--22937, 2019.

\bibitem[Cui et~al.(2019)Cui, Bucur, Groot, and Heskes]{cui2019novel}
Ruifei Cui, Ioan~Gabriel Bucur, Perry Groot, and Tom Heskes.
\newblock A novel bayesian approach for latent variable modeling from mixed
  data with missing values.
\newblock \emph{Statistics and Computing}, 29\penalty0 (5):\penalty0 977--993,
  2019.

\bibitem[Davenport et~al.(2014)Davenport, Plan, Van Den~Berg, and
  Wootters]{davenport20141}
Mark~A Davenport, Yaniv Plan, Ewout Van Den~Berg, and Mary Wootters.
\newblock 1-bit matrix completion.
\newblock \emph{Information and Inference: A Journal of the IMA}, 3\penalty0
  (3):\penalty0 189--223, 2014.

\bibitem[Fan et~al.(2017)Fan, Liu, Ning, and Zou]{fan2017high}
Jianqing Fan, Han Liu, Yang Ning, and Hui Zou.
\newblock High dimensional semiparametric latent graphical model for mixed
  data.
\newblock \emph{Journal of the Royal Statistical Society: Series B (Statistical
  Methodology)}, 79\penalty0 (2):\penalty0 405--421, 2017.

\bibitem[Fan and Udell(2019)]{fan2019online}
Jicong Fan and Madeleine Udell.
\newblock Online high rank matrix completion.
\newblock In \emph{Proceedings of the IEEE Conference on Computer Vision and
  Pattern Recognition}, pages 8690--8698, 2019.

\bibitem[Fan et~al.(2020)Fan, Zhang, and Udell]{fan2020polynomial}
Jicong Fan, Yuqian Zhang, and Madeleine Udell.
\newblock Polynomial matrix completion for missing data imputation and
  transductive learning.
\newblock In \emph{Thirty-Fourth AAAI Conference on Artificial Intelligence},
  pages 3842--3849, 2020.

\bibitem[Feng and Ning(2019)]{feng2019high}
Huijie Feng and Yang Ning.
\newblock High-dimensional mixed graphical model with ordinal data: Parameter
  estimation and statistical inference.
\newblock In \emph{The 22nd International Conference on Artificial Intelligence
  and Statistics}, pages 654--663, 2019.

\bibitem[Ganti et~al.(2015)Ganti, Balzano, and Willett]{ganti2015matrix}
Ravi~Sastry Ganti, Laura Balzano, and Rebecca Willett.
\newblock Matrix completion under monotonic single index models.
\newblock In \emph{Advances in Neural Information Processing Systems}, pages
  1873--1881, 2015.

\bibitem[Goldberg et~al.(2010)Goldberg, Recht, Xu, Nowak, and
  Zhu]{goldberg2010transduction}
Andrew Goldberg, Ben Recht, Junming Xu, Robert Nowak, and Jerry Zhu.
\newblock Transduction with matrix completion: Three birds with one stone.
\newblock In \emph{Advances in Neural Information Processing Systems}, pages
  757--765, 2010.

\bibitem[Gunasekar et~al.(2014)Gunasekar, Ravikumar, and
  Ghosh]{gunasekar2014exponential}
Suriya Gunasekar, Pradeep Ravikumar, and Joydeep Ghosh.
\newblock Exponential family matrix completion under structural constraints.
\newblock In \emph{International Conference on Machine Learning}, pages
  1917--1925, 2014.

\bibitem[Guo et~al.(2015)Guo, Levina, Michailidis, and Zhu]{guo2015graphical}
Jian Guo, Elizaveta Levina, George Michailidis, and Ji~Zhu.
\newblock Graphical models for ordinal data.
\newblock \emph{Journal of Computational and Graphical Statistics}, 24\penalty0
  (1):\penalty0 183--204, 2015.

\bibitem[Harper and Konstan(2015)]{harper2015movielens}
F~Maxwell Harper and Joseph~A Konstan.
\newblock The movielens datasets: History and context.
\newblock \emph{ACM Transactions on Interactive Intelligent Systems},
  5\penalty0 (4):\penalty0 19, 2015.

\bibitem[Hastie and Mazumder(2015)]{hastie2015softimpute}
T~Hastie and R~Mazumder.
\newblock softimpute: Matrix completion via iterative soft-thresholded svd.
\newblock \emph{R package version}, 1, 2015.

\bibitem[Hoff et~al.(2007)]{hoff2007extending}
Peter~D Hoff et~al.
\newblock Extending the rank likelihood for semiparametric copula estimation.
\newblock \emph{The Annals of Applied Statistics}, 1\penalty0 (1):\penalty0
  265--283, 2007.

\bibitem[Hsu et~al.(2012)Hsu, Kakade, Zhang, et~al.]{hsu2012tail}
Daniel Hsu, Sham Kakade, Tong Zhang, et~al.
\newblock A tail inequality for quadratic forms of subgaussian random vectors.
\newblock \emph{Electronic Communications in Probability}, 17, 2012.

\bibitem[Ilin and Raiko(2010)]{ilin2010practical}
Alexander Ilin and Tapani Raiko.
\newblock Practical approaches to principal component analysis in the presence
  of missing values.
\newblock \emph{Journal of Machine Learning Research}, 11\penalty0
  (Jul):\penalty0 1957--2000, 2010.

\bibitem[Josse et~al.(2011)Josse, Pag{\`e}s, and Husson]{josse2011multiple}
Julie Josse, J{\'e}r{\^o}me Pag{\`e}s, and Fran{\c{c}}ois Husson.
\newblock Multiple imputation in principal component analysis.
\newblock \emph{Advances in data analysis and classification}, 5\penalty0
  (3):\penalty0 231--246, 2011.

\bibitem[Josse et~al.(2016)Josse, Husson, et~al.]{josse2016missmda}
Julie Josse, Fran{\c{c}}ois Husson, et~al.
\newblock missmda: a package for handling missing values in multivariate data
  analysis.
\newblock \emph{Journal of Statistical Software}, 70\penalty0 (1):\penalty0
  1--31, 2016.

\bibitem[Keshavan et~al.(2010)Keshavan, Montanari, and Oh]{keshavan2010matrix}
Raghunandan~H Keshavan, Andrea Montanari, and Sewoong Oh.
\newblock Matrix completion from noisy entries.
\newblock \emph{Journal of Machine Learning Research}, 11\penalty0
  (Jul):\penalty0 2057--2078, 2010.

\bibitem[Lan et~al.(2014)Lan, Studer, and Baraniuk]{lan2014matrix}
Andrew~S Lan, Christoph Studer, and Richard~G Baraniuk.
\newblock Matrix recovery from quantized and corrupted measurements.
\newblock In \emph{2014 IEEE International Conference on Acoustics, Speech and
  Signal Processing}, pages 4973--4977. IEEE, 2014.

\bibitem[Lim and Teh(2007)]{lim2007variational}
Yew~Jin Lim and Yee~Whye Teh.
\newblock Variational bayesian approach to movie rating prediction.
\newblock In \emph{Proceedings of KDD cup and workshop}, volume~7, pages
  15--21. Citeseer, 2007.

\bibitem[Little and Rubin(2002)]{little2002statistical}
Roderick~JA Little and Donald~B Rubin.
\newblock \emph{Statistical analysis with missing data}.
\newblock Wiley, 2002.

\bibitem[Liu et~al.(2009)Liu, Lafferty, and Wasserman]{liu2009nonparanormal}
Han Liu, John Lafferty, and Larry Wasserman.
\newblock The nonparanormal: Semiparametric estimation of high dimensional
  undirected graphs.
\newblock \emph{Journal of Machine Learning Research}, 10\penalty0
  (Oct):\penalty0 2295--2328, 2009.

\bibitem[Mazumder et~al.(2010)Mazumder, Hastie, and
  Tibshirani]{mazumder2010spectral}
Rahul Mazumder, Trevor Hastie, and Robert Tibshirani.
\newblock Spectral regularization algorithms for learning large incomplete
  matrices.
\newblock \emph{Journal of machine learning research}, 11\penalty0
  (Aug):\penalty0 2287--2322, 2010.

\bibitem[McLachlan and Krishnan(2007)]{mclachlan2007algorithm}
Geoffrey~J McLachlan and Thriyambakam Krishnan.
\newblock \emph{The EM algorithm and extensions}, volume 382.
\newblock John Wiley \& Sons, 2007.

\bibitem[Mnih and Salakhutdinov(2008)]{mnih2008probabilistic}
Andriy Mnih and Russ~R Salakhutdinov.
\newblock Probabilistic matrix factorization.
\newblock In \emph{Advances in neural information processing systems}, pages
  1257--1264, 2008.

\bibitem[Murray et~al.(2013)Murray, Dunson, Carin, and
  Lucas]{murray2013bayesian}
Jared~S Murray, David~B Dunson, Lawrence Carin, and Joseph~E Lucas.
\newblock Bayesian gaussian copula factor models for mixed data.
\newblock \emph{Journal of the American Statistical Association}, 108\penalty0
  (502):\penalty0 656--665, 2013.

\bibitem[Ongie et~al.(2017)Ongie, Willett, Nowak, and
  Balzano]{ongie2017algebraic}
Greg Ongie, Rebecca Willett, Robert~D Nowak, and Laura Balzano.
\newblock Algebraic variety models for high-rank matrix completion.
\newblock In \emph{Proceedings of the 34th International Conference on Machine
  Learning-Volume 70}, pages 2691--2700. JMLR. org, 2017.

\bibitem[Pakman and Paninski(2014)]{pakman2014exact}
Ari Pakman and Liam Paninski.
\newblock Exact hamiltonian monte carlo for truncated multivariate gaussians.
\newblock \emph{Journal of Computational and Graphical Statistics}, 23\penalty0
  (2):\penalty0 518--542, 2014.

\bibitem[Recht et~al.(2010)Recht, Fazel, and Parrilo]{recht2010guaranteed}
Benjamin Recht, Maryam Fazel, and Pablo~A Parrilo.
\newblock Guaranteed minimum-rank solutions of linear matrix equations via
  nuclear norm minimization.
\newblock \emph{SIAM review}, 52\penalty0 (3):\penalty0 471--501, 2010.

\bibitem[Rennie and Srebro(2005)]{rennie2005fast}
Jasson~DM Rennie and Nathan Srebro.
\newblock Fast maximum margin matrix factorization for collaborative
  prediction.
\newblock In \emph{Proceedings of the 22nd International Conference on Machine
  learning}, pages 713--719. ACM, 2005.

\bibitem[Robin et~al.(2020)Robin, Klopp, Josse, Moulines, and
  Tibshirani]{robin2020main}
Genevi{\`e}ve Robin, Olga Klopp, Julie Josse, {\'E}ric Moulines, and Robert
  Tibshirani.
\newblock Main effects and interactions in mixed and incomplete data frames.
\newblock \emph{Journal of the American Statistical Association}, 115\penalty0
  (531):\penalty0 1292--1303, 2020.

\bibitem[Rubin(1996)]{rubin1996multiple}
Donald~B Rubin.
\newblock Multiple imputation after 18+ years.
\newblock \emph{Journal of the American statistical Association}, 91\penalty0
  (434):\penalty0 473--489, 1996.

\bibitem[Salakhutdinov and Mnih(2008)]{salakhutdinov2008bayesian}
Ruslan Salakhutdinov and Andriy Mnih.
\newblock Bayesian probabilistic matrix factorization using markov chain monte
  carlo.
\newblock In \emph{Proceedings of the 25th international conference on Machine
  learning}, pages 880--887, 2008.

\bibitem[Tipping and Bishop(1999)]{tipping1999probabilistic}
Michael~E Tipping and Christopher~M Bishop.
\newblock Probabilistic principal component analysis.
\newblock \emph{Journal of the Royal Statistical Society: Series B (Statistical
  Methodology)}, 61\penalty0 (3):\penalty0 611--622, 1999.

\bibitem[Udell and Townsend(2019)]{udell2019why}
Madeleine Udell and Alex Townsend.
\newblock Why are big data matrices approximately low rank?
\newblock \emph{SIAM Journal on Mathematics of Data Science (SIMODS)},
  1\penalty0 (1):\penalty0 144--160, 2019.

\bibitem[Udell et~al.(2016)Udell, Horn, Zadeh, Boyd,
  et~al.]{udell2016generalized}
Madeleine Udell, Corinne Horn, Reza Zadeh, Stephen Boyd, et~al.
\newblock Generalized low rank models.
\newblock \emph{Foundations and Trends{\textregistered} in Machine Learning},
  9\penalty0 (1):\penalty0 1--118, 2016.

\bibitem[Vershynin(2010)]{vershynin2010introduction}
Roman Vershynin.
\newblock Introduction to the non-asymptotic analysis of random matrices.
\newblock \emph{arXiv preprint arXiv:1011.3027}, 2010.

\bibitem[Yang et~al.(2019)Yang, Akimoto, Kim, and Udell]{yang2019oboe}
Chengrun Yang, Yuji Akimoto, Dae~Won Kim, and Madeleine Udell.
\newblock Oboe: Collaborative filtering for automl model selection.
\newblock In \emph{Proceedings of the 25th ACM SIGKDD International Conference
  on Knowledge Discovery \& Data Mining}, pages 1173--1183, 2019.

\bibitem[Yu et~al.(2009)Yu, Zhu, Lafferty, and Gong]{yu2009fast}
Kai Yu, Shenghuo Zhu, John Lafferty, and Yihong Gong.
\newblock Fast nonparametric matrix factorization for large-scale collaborative
  filtering.
\newblock In \emph{Proceedings of the 32nd ACM SIGIR International Conference
  on Research and Development in Information Retrieval}, pages 211--218, 2009.

\bibitem[Zhao and Udell(2020)]{zhao2020missing}
Yuxuan Zhao and Madeleine Udell.
\newblock Missing value imputation for mixed data via gaussian copula.
\newblock In \emph{Proceedings of the 26th ACM SIGKDD International Conference
  on Knowledge Discovery \& Data Mining}, pages 636--646, 2020.

\end{thebibliography}

\begin{appendix}
\clearpage

\section{Proofs}

\paragraph{Setup}
Suppose a $p$-dimensional vector $\bx \sim\lrgc$ is observed at locations $\indexO\subset [p]$ and missing at $\indexM=[p]/\indexO$.
Then according to the definition of LRGC,
for $\bt\sim \mathcal{N}(\bo,\mathrm{I}_k)$, $\beps \sim \mathcal{N}(\bo,\sigma^2\mathrm{I}_p)$, and $\bz = \bW \bt + \beps$,
we know $\bx = \bigG(\bz)$ and
$\bz\sim \mathcal{N}(\bo, \mathbf{\Sigma})$ with $\mathbf{\Sigma}=\bW\bW^\intercal + \sigma^2 \Irm_p$.
Here we say two random vectors are equal if they have the same CDF.

A key fact we use is that conditional on known $\zobs$, $\zmis$ has a normal distribution:
\begin{equation}
	\zmis|\zobs  \sim \mathcal{N}(\mathbf{\Sigma}_{\indMis,\indObs}\mathbf{\Sigma}_{\indObs,\indObs}^{-1}\zobs,\mathbf{\Sigma}_{\indexM,\indexM} - \mathbf{\Sigma}_{\indMis,\indObs}\mathbf{\Sigma}_{\indObs,\indObs}^{-1}\mathbf{\Sigma}_{\indObs,\indexM}).
	\label{Eq:conditiona_distribution}
\end{equation}
Here we use $\mathbf{\Sigma}_{I, J}$ to denote the submatrix of $\mathbf{\Sigma}$ with rows in $I$ and columns in $J$.
Plugging in $\mathbf{\Sigma}=\bW\bW^\intercal + \sigma^2 \Irm_p$, 
we obtain 
\begin{align}
	\Erm[\zmis|\zobs]& = \Wmis\Wobs^\intercal (\Wobs\Wobs^\intercal + \sigma^2\Irm)^{-1}\zobs 
	\nonumber\\
	&=\Wmis (\sigma^2\Irm + \Wobs^\intercal \Wobs )^{-1}\Wobs ^\intercal \zobs.
	\label{Eq:first_moment}
\end{align}
In last equation, we use the Woodbury matrix identity.
Similarly, we obtain:
\begin{equation}
	\Covrm[\zmis|\zobs] = \sigma^2 \Irm + \sigma^2 \Wmis (\sigma^2\Irm + \Wobs^\intercal \Wobs )^{-1}\Wmis^\intercal.
	\label{Eq:second_moment}
\end{equation}

\subsection{Proof for Lemma 1}
\label{section:lemma1}
Using the law of total expectation,
\begin{align*}
    \Erm[\zmis|\xobs] &= \Erm\left[\Erm[\zmis|\zobs]\big |\xobs\right]
    \nonumber\\
    &= \Erm[\Wmis (\sigma^2\Irm + \Wobs^\intercal \Wobs )^{-1}\Wobs ^\intercal \zobs|\xobs]
    \nonumber\\
    &=\Wmis (\sigma^2\Irm + \Wobs^\intercal \Wobs )^{-1}\Wobs ^\intercal \Erm[\zobs|\xobs].
    \nonumber
\end{align*}
For the first equality,
we use \cref{Eq:first_moment}.

Similarly we can compute the second moments,
\begin{align}
    \Erm[\zmis \zmis^\intercal|\xobs] &= \Erm\left[\Erm[\zmis \zmis^\intercal|\zobs]\big |\xobs\right]\nonumber\\
    &= \Erm\left[\Erm[\zmis|\zobs] \Erm[\zmis|\zobs]^\intercal + \Covrm[\zmis|\zobs]|\xobs\right]
    \nonumber\\
    &=\Erm\left[\Erm[\zmis|\zobs] \Erm[\zmis|\zobs]^\intercal |\xobs\right] + \Erm\left[\Covrm[\zmis|\zobs]|\xobs\right] 
    \nonumber\\
    &=\Erm\left[\Erm[\zmis|\zobs] \Erm[\zmis|\zobs]^\intercal |\xobs\right] + \Covrm[\zmis|\zobs].
    \label{Eq:lemma1_step2}
\end{align}
From the last equation, we use the fact that  $\Covrm[\zmis|\xobs]$ is fully determined by $\bW$ and $\sigma^2$ and thus does not depend on $\xmis$.

Plug \cref{Eq:first_moment} and \cref{Eq:second_moment} into \cref{Eq:lemma1_step2}
to obtain 
\begin{multline*}
        \Erm\left[\Erm[\zmis|\zobs] \Erm[\zmis|\zobs]^\intercal |\xobs\right] =\\
        \Wmis (\sigma^2\Irm + \Wobs^\intercal \Wobs )^{-1}\Wobs ^\intercal \Erm[\zobs \zobs^\intercal|\xobs] \Wobs  (\sigma^2\Irm + \Wobs^\intercal \Wobs )^{-1}\Wmis ^\intercal. 
\end{multline*}
Then using $\Covrm[\zmis|\xobs]=\Erm[\zmis \zmis^\intercal|\xobs]-\Erm[\zmis|\xobs]\Erm[\zmis^\intercal|\xobs]$, we have
\begin{multline*}
    \Covrm[\zmis|\xobs]=\sigma^2\Irm_{|\indexM|} +\sigma^2 \Wmis(\sigma^2\Irm + \Wobs^\intercal \Wobs )^{-1}\Wmis^\intercal + \\		
			 \Wmis (\sigma^2\Irm + \Wobs^\intercal \Wobs )^{-1}\Wobs^\intercal\Covrm[\zobs|\xobs]\Wobs (\sigma^2\Irm + \Wobs^\intercal \Wobs )^{-1}\Wmis^\intercal.
\end{multline*}

\subsection{Proof of Theorem 1}
\begin{proof}
Theorem 1 is an immediate consequence of the normality of $\zmis$ conditional on $\zobs=\bigG_{\indexO}^{-1}(\xobs)$ (see \cref{Eq:conditiona_distribution}) and the elementwise strictly monotone $\bigG$.
\end{proof}

\subsection{Proof of Theorem 2}
Suppose $\bx=(x_1,\ldots,x_p)$ where $x_j$ is ordinal with $k_j(\geq 2)$ ordinal levels encoded as $\{1,\ldots,k_j\}$ for $j\in[p]$.
For ordinal data, the conditional distribution of $\zmis|\zobs \in \bigG_{\indexO}^{-1}(\xobs)$ is intractable. 
Consequently, we cannot establish distribution-based confidence intervals for $\zmis$.

Instead, for each marginal $j$,
we can lower bound the probability of event $|\hat x_j -x_j|\leq d$ for the LRGC imputation $\hat x_j$ and $d\in \mathbb{Z}$.
Since $\Pr(|\hat x_j-x_j|\leq k_j-1)=1$, it suffices to consider $d<k_j-1$.
In practice, the result is more useful for small $d$, such as $d=0$.
Let us first state a generalization of our Theorem 2.

\setcounter{theorem}{3}

\begin{theorem}
	Suppose $\bx \sim \lrgc$ with observations $\xobs$ and missing entries $\xmis$.
	Also for each marginal $j\in[p]$,
	$x_j$ takes values from $\{1,\ldots,k_j\}$ 
	and thus the $g_j$ is a step function with cut points $\bS_j=\{s_1,\ldots,s_{k_j-1}\}$:
	\begin{equation*}
	        g_j(z) = 1 + \sum_{k=1}^{k_j-1}\mathds{1}(z>s_k), 
	        \quad \mbox{ where }-\infty=:s_0<s_1<\ldots<s_{k_j-1}<s_{k_j}:=\infty.
	\end{equation*}
	For a missing entry $x_j$, $j \in \mathcal M$,
	the set of values for $z_j$ that would yield the same imputed value $\hat x_j=g_j(\Erm[z_j|\xobs])$ is $g_j^{-1}(\hat x_j)=(s_{\hat x_j-1},s_{\hat x_j}]$.
	Then the following holds:
	\begin{equation*}
	\Pr(|\hat x_j-x_j|\leq d)\geq 1 -\frac{\Varrm[z_j|\xobs]}{d^2_{j}},
	\end{equation*}
	with
	\[
		d_{j}=\min(|\Erm[z_j|\xobs]-s_{\max(\hat x_j-1-d,0)}|, |\Erm[z_j|\xobs]-s_{\min(\hat x_j+d,k_j)}|),
	\]
	where $\Erm[z_j|\xobs],\Varrm[z_j|\xobs]$ are given in Lemma 1 with $\indexM$ replaced by $j$.
	\label{theorem:uncertainty_ordinal_general}
\end{theorem}

\begin{proof}
The proof applies to each missing dimension $j\in\indexM$.
Let us further define $s_{k}=-\infty$ for any negative integer $k$ 
and $s_{k}=\infty$ for any integer $k>k_j$ for convenience.
Then $s_k = s_{\max(k,0)}$ for negative integer $k$ and $s_k = s_{\min(k,k_j)}$ for integer $k$ larger than $k_j$.

First notice $|\hat x_j -x_j|\leq d$ if and only if $z_j\in(s_{\hat x_j-1-d},s_{\hat x_j+d}]$ for the latent normal $z_j$ satisfying $x_j=g_j(z_j)$.
Specifically, when $d=0$, $\hat x_j =x_j$ if and only if
$z_j\in (s_{\hat x_j-1},s_{\hat x_j}]$,
i.e. $g_j^{-1}( x_j)=(s_{\hat x_j-1},s_{\hat x_j}]=g_j^{-1}(\hat x_j)$.
Notice we have,
 \[
 \Erm[z_j|\xobs]\in (s_{\hat x_j-1},s_{\hat x_j}] \subset (s_{\hat x_j-1-d},s_{\hat x_j+d}].
 \]
Thus a sufficient condition for $z_j\in (s_{\hat x_j-1-d},s_{\hat x_j+d}]$ is that  $z_j$ is sufficiently close to its conditional mean $\Erm[z_j|\xobs]$.
More precisely, 
\begin{equation*}
    |\Erm[z_j|\xobs]-z_j|\leq \min(|\Erm[z_j|\xobs]-s_{\hat x_j-1-d}|, |\Erm[z_j|\xobs]-s_{\hat x_j+d}|) \rightarrow |\hat x_j - x_j|\leq d.
    \label{Eq:sufficient_condtion}
\end{equation*}
 
Define $d_j:=\min(|\Erm[z_j|\xobs]-s_{\hat x_j-1-d}|, |\Erm[z_j|\xobs]-s_{\hat x_j+d}|)$. 
Notice when $d=0$, 
\[
d_j=\min(|\Erm[z_j|\xobs]-s_{\hat x_j-1}|, |\Erm[z_j|\xobs]-s_{\hat x_j}|)=\min_{s\in \bS}|\Erm[z_j|\xobs]-s|.
\]
Use the Markov inequality together with the conditional distribution of $z_j$ given $\xobs$ to bound
 \begin{equation*}
     \Pr( |\Erm[z_j|\xobs]-z_j|>d_j) \leq \frac{\Varrm[z_j|\xobs]}{d_j^2},
 \end{equation*}
 which completes our proof.
\end{proof}


\subsection{Proof of Theorem 3}
To prove Theorem 3, we introduce a lemma which provides a concentration inequality on quadratic forms of sub-Gaussian vectors.
For a detailed treatment of sub-Gaussian random distributions, see \cite{vershynin2010introduction}.
A random variable $x\in \Rbb$ is called sub-Gaussian if $(\Erm[|x|^p])^{1/p}\leq K\sqrt{p}$ for all $p\geq 1$ with some $K>0$.
The sub-Gaussian norm of $x$ is defined as $||x||_{\psi_2}=\sup_{p\geq 1}p^{-1/2}(\Erm[|x|^p])^{1/p}$.

Denote  the inner product of vectors $\bx_1$ and $\bx_2$ as $\langle \bx_1,\bx_2\rangle$.
A random vector $\bx\in \Rbb^p$ is called sub-Gaussian if the one-dimensional marginals $\langle \bx ,\mathbf{a}\rangle$ are all sub-Gaussian random variables for any constant vector $\mathbf{a}\in \Rbb^p$.
The sub-Gaussian norm of $\bx$ is defined as  $||\bx||_{\psi_2}=\sup_{\mathbf{a}\in \mathbb{S}^{p-1}}||\langle \bx,\mathbf{a} \rangle||_{\psi_2}$.
A Gaussian random vector is also sub-Gaussian.

\setcounter{lemma}{1}
\begin{lemma}
	Let $\Sigma\in \Rbb^{p\times p}$ be a positive semidefinite matrix.
	 Let $\bx = (x_1,\ldots,x_p)$ be a sub-Gaussian random vector with mean zero and covariance matrix $\Irm_p$. For all $t>0$,
	\[
	\Pr\left[\bx^\intercal \Sigma \bx > (\sqrt{\tr(\Sigma)}+\sqrt{2\lammax(\Sigma)t})^2\right]\leq e^{-t}.
	\]
	\label{lemma:subgaussian}
\end{lemma}
Our \cref{lemma:subgaussian} is Lemma 17 in \cite{bing2020adaptive}, which is also a simplified version of Theorem 1 in \cite{hsu2012tail}.

\begin{proof}
Since $\bigG$ is elementwise Lipschitz with constant $L$,
\begin{equation}
\mbox{MSE}(\hat \bx) = \frac{||\bigG_{\indexM}(\zmis)-\bigG_{\indexM}(\hat \bz_{\indexM})||_2^2}{||\indexM||}\leq L^2 \frac{||\zmis-\hat \bz_{\indexM}||_2^2}{||\indexM||}.
\label{Eq:lipschitz}
\end{equation}

Denote the covariance matrix of $\zmis$ conditional on $\zobs$ as $\mathbf{\Sigma}_{(\indexM)}$.
Apply the above inequality with $\mathbf{\Sigma}=\mathbf{\Sigma}_{(\indexM)}$ and $\bx = \mathbf{\Sigma}_{(\indexM)}^{-1/2}\zmis$,
 we obtain: 
 \begin{equation}
 	\Pr\left(||\zmis-\hat \bz_{\indexM}||_2^2 > \left(\sqrt{\tr(\mathbf{\Sigma}_{(\indexM)})}+\sqrt{2\lammax(\mathbf{\Sigma}_{(\indexM)})t}\right)^2\right)\leq e^{-t}.
 	\label{Eq:bound_z_quadratic_1}
 \end{equation}
Notice
\begin{align*}
	\tr(\Sigma_{(\indexM)})&=\tr\left(\sigma^2 \Irm + \sigma^2 \Wmis (\sigma^2\Irm + \Wobs^\intercal \Wobs )^{-1}\Wmis^\intercal\right)
	\nonumber\\
	&=\sigma^2|\indexM|+\sigma^2 \tr\left( (\sigma^2\Irm + \Wobs^\intercal \Wobs )^{-1}\Wmis^\intercal\Wmis\right)
	\nonumber\\
	&\leq \sigma^2|\indexM|+\sigma^2 \lammax(\sigma^2\Irm + \Wobs^\intercal \Wobs )^{-1})\tr\left(\Wmis^\intercal\Wmis\right)
	\label{Eq:tr_step2}\\
	&=  \sigma^2|\indexM|+\sigma^2  \frac{1}{\sigma^2+\lammin^2(\Wobs)} (1-\sigma^2)|\indexM|.
	\nonumber
\end{align*}
In the inequality, we use the fact $\tr(\mathbf{AB})\leq \lammax(\mathbf{A})\tr(\mathbf{B})$ for any  real symmetric positive semidefinite matrices $\mathbf{A}$ and $\mathbf{B}$.
In the last equation, we use the unit diagonal constraints of $\bW\bW^\intercal +\sigma^2\Irm_p$ such that $\tr(\Wmis^\intercal\Wmis)=\tr(\Wmis \Wmis^\intercal)=|\indexM|(1-\sigma^2)$.

Also notice
\begin{align*}
	\lammax(\Sigma_{(\indexM)})&=\lammax( \sigma^2 \Irm + \sigma^2 \Wmis (\sigma^2\Irm + \Wobs^\intercal \Wobs )^{-1}\Wmis^\intercal)\\
	&\leq \sigma^2 + \sigma^2 \lammax(\Wmis (\sigma^2\Irm + \Wobs^\intercal \Wobs )^{-1}\Wmis^\intercal) \\
	& \leq \sigma^2 + \sigma^2 \lammax^2(\Wmis) \lammax((\sigma^2\Irm + \Wobs^\intercal \Wobs )^{-1}) \\
	&=\sigma^2+\sigma^2 \frac{\lammax^2(\Wmis)}{\sigma^2+\lammin^2(\Wobs)}.
\end{align*}
Thus,
\begin{equation}
	||\zmis-\hat \bz_{\indexM}||_2^2 \leq \sigma^2 |\indexM|\cdot \left(\sqrt{1+\frac{1-\sigma^2}{\sigma^2+\lammin^2(\Wobs)}}+\sqrt{\left(1+\frac{\lammax^2(\Wmis)}{\sigma^2+\lammin^2(\Wobs)}\right)\frac{2t}{|\indexM|}}\right)^2.
	\label{Eq:bound_z_quadratic_2}
\end{equation}
Combining \cref{Eq:lipschitz}, \cref{Eq:bound_z_quadratic_1} and \cref{Eq:bound_z_quadratic_2}, we finish the proof.
\end{proof}

\subsection{Proof of Corollary 1}
We first introduce a result from \cite[Theorem~5.39]{vershynin2010introduction} characterizing the singular values of long random matrices with independent sub-Gaussian rows.
\begin{lemma}
	Let $\mathbf{A}\in \Rbb^{p\times k}$ be a matrix whose rows $\mathbf{a}_j$ are independent sub-Gaussian random vectors in $\Rbb^k$ whose covariance matrix is $\mathbf{\Sigma}$. Then for every $t>0$, with probability as least $1-2\exp(-ct^2)$ one has 
	\begin{equation*}
			\lammax\left(\frac{1}{p}\mathbf{A}^\intercal \mathbf{A} - \mathbf{\Sigma}\right)\leq \max(\delta, \delta^2)\lammax(\mathbf{\Sigma}), \quad \mbox{where }\delta = C\sqrt{\frac{k}{p}}+\frac{t}{\sqrt{p}}.
	\end{equation*}
	Here $c,\,C>0$ depend only on the subgaussian norm $K=\max_j||\Sigma^{-1/2}\mathbf{a}_j||_{\psi_2}$.
	\label{lemma:singular_value}
\end{lemma}

\begin{proof}

Apply \cref{lemma:singular_value} to submatrix $\Wobs$ and $\Wmis$ respectively with covariance matrix $\mathbf{\Sigma}=\frac{1-\sigma^2}{k}\Irm_k$,
we obtain with probability at least $1-2\exp(-ct_1^2)-2\exp(-ct_2^2)$,
\begin{equation*}
	\left|\frac{1}{|\indexO|}\lammin^2(\Wobs) - \frac{1-\sigma^2}{k}\right|\leq  \frac{1-\sigma^2}{k}\epsilon_1 \mbox{ and }	\left|\frac{1}{|\indexM|}\lammax^2(\Wmis) - \frac{1-\sigma^2}{k}\right|\leq  \frac{(1-\sigma^2)\epsilon_2}{k},
\end{equation*}
where $\epsilon_1=\max(\delta_1,\delta^2_1)$ with $\delta_1=\frac{C\sqrt{k}+t_1}{\sqrt{|\indexO|}}$ and 
$\epsilon_2=\max(\delta_2,\delta^2_2)$ with $\delta_2=\frac{C\sqrt{k}+t_2}{\sqrt{|\indexM|}}$.
Constants  $c,\,C>0$ only depend on the  subgaussian norm $\max_j||\sqrt{\frac{k}{1-\sigma^2}}\bw_j||_{\psi_2}$.

For any $0<\epsilon<1$, let $t_1=\frac{\epsilon \sqrt{|\indexO|}}{2}$ and $t_2=\frac{\epsilon\sqrt{|\indexO|}}{2\sqrt{c_2}}$.
 Suppose the sufficiently large constant $c_1$ satisfies $c_1>\frac{4C^2\max(1,c_2)}{\epsilon^2}$.
Then we have 
\begin{align*}
    \epsilon_1&=\delta_1= \frac{C}{\sqrt{|\indexO|/k}} + \frac{t}{\sqrt{|\indexO|}}< \frac{C}{\sqrt{c_1}} + \frac{\epsilon}{2}<
    \frac{C}{\sqrt{4C^2/\epsilon^2}} + \frac{\epsilon}{2}=\epsilon,
\end{align*}
and 
\[
    \epsilon_2=\delta_2= \frac{C}{\sqrt{|\indexM|/k}} + \frac{t}{\sqrt{|\indexM|}}< \frac{C}{\sqrt{|\indexO|/c_2k}} + \frac{\epsilon}{2}<
    \frac{C}{\sqrt{4C^2/\epsilon^2}} + \frac{\epsilon}{2}=\epsilon.
\]
 
 Thus we have with probability at least $1-2\exp(-{c\epsilon^2|\indexO|}/{4})-2\exp(-{c\epsilon^2|\indexO|}/{4c_2})$,
 \begin{equation}
 	\lammin^2(\Wobs) > (1-\sigma^2)(1-\epsilon)\frac{|\indexO|}{k}\quad \mbox{ and }\quad \lammax^2(\Wmis) \leq (1-\sigma^2)(1+\epsilon)\frac{|\indexM|}{k}.
 	\label{Eq:bound_singular_value}
 \end{equation}
 Combining \cref{Eq:bound_z_quadratic_2} and \cref{Eq:bound_singular_value},
 then with probability at least $1-\exp(-t)-2\exp({-c\epsilon^2|\indexO|}/{4})-2\exp({-c\epsilon^2|\indexO|}/{4c_2})$,
 \begin{align}
      	\frac{||\zmis-\hat \bz_{\indexM}||_2^2}{|\indexM|}
      	&\leq \sigma^2 \left(\sqrt{1+\frac{1}{\frac{\sigma^2}{1-\sigma^2}+(1-\epsilon)|\indexO|/k}}+
 	\sqrt{\frac{2t}{|\indexM|}+
 		\frac{2(1+\epsilon)t}{\frac{k\sigma^2}{1-\sigma^2}+(1-\epsilon)|\indexO|}}\right)^2
 		\nonumber\\
 		&\leq \sigma^2 \left(\sqrt{1+\frac{1}{\frac{\sigma^2}{1-\sigma^2}+(1-\epsilon)|\indexO|/k}}+
 	\sqrt{\frac{2c_2t}{|\indexO|}+
 		\frac{2(1+\epsilon)t}{\frac{k\sigma^2}{1-\sigma^2}+(1-\epsilon)|\indexO|}}\right)^2.
 		\label{Eq:zw_upper_bound}
 \end{align}
Now take $t=\log |\indexO|$,
with fixed $k$ and $\sigma^2$,
the right hand side is $1+O\left(\sqrt{\frac{\log |\indexO|}{|\indexO|}}\right)$.

Notice $|\indexO|>c_1k\geq c_1$. 
Then there exists some constant $c_3>0$ such that $|\indexO|$ satisfies:
\begin{equation*}
    \log |\indexO| < c_3\frac{c\epsilon^2|\indexO|}{4\max(1,c_2)}
\end{equation*}
thus \cref{Eq:zw_upper_bound} holds with probability at least $1-\frac{1+2c_3}{|\indexO|}$.
Combing the result with \cref{Eq:lipschitz} completes the proof.
\end{proof}

\section{Algorithm detail}
\subsection{E-step details}
We provide details on E-step computation here.
The key fact we use is that conditional on known $\zobs$, $\bt$ is normally distributed:
\begin{equation}
\bt|\bz_{\indexO} \sim \mathcal{N}(\bM^{-1}_{\indexO}\Wobs^\intercal\bz_{\indexO}, \sigma^2\bM_{\indexO}^{-1}),
\end{equation}
where $\bM_{\indexO}=\sigma^2 \Irm_k + \Wobs^\intercal  \Wobs$. 
This result follows by applying the Bayes formula with $\zobs|\bt \sim \mathcal{N}(\Wobs \bt,\sigma^2\Irm_p), \bt \sim \mathcal{N}(\bo,\Irm_k)$ and $\zobs \sim \mathcal{N}(\bo,\Wobs\Wobs^\intercal +\sigma^2\Irm_p)$.

First we express the Q-function $Q(Q(\bW,\sigma^2;\tilde \bW, \tilde \sigma^2))$:
\begin{multline}
    Q(\bW,\sigma^2;\tilde \bW, \tilde \sigma^2)=
    c-\frac{\sum_{i=1}^n|\indexO_i|\log(\sigma^2)}{2}\\
    -\frac{\sum_{i=1}^n\left(
    	\Ebb[(\zio)^\intercal \zio] - 2\tr(\Wiobs \Ebb[\bt^i (\zio)^\intercal]) + \tr(\Wiobs^\intercal \Wiobs \Ebb[\bt^i(\bt^i)^\intercal])
    	\right)}{2\sigma^2},
    	\label{Eq;Q-func}
\end{multline}
where $c$ is an absolute constant in terms the model parameters $\bW$ and $\sigma^2$.

Thus to evaluate the $Q$ function, 
we only need (1)$\Ebb[(\zio)^\intercal \zio]$,
(2)$\Ebb[\bt^i(\zio)^\intercal]$ and 
(3)$\Ebb[\bt^i(\bt^i)^\intercal]$.
computing (1) only needs $\Erm[\zio|\xio]$ and $\Covrm[\zio|\xio]$. 
To compute (2) and (3),
we use the law of total expectation similar as in \cref{section:lemma1} by first treating $\zio$ as known.

Since $\Erm[\bt^i|\zio]=\bM^{-1}_{\indexO_i}\Wiobs^\intercal\zio$
and 
$\Covrm[\bt^i|\zio]=\sigma^2\bM^{-1}_{\indexO_i}$,
we have
\begin{align*}
    \Erm[\bt^i|\xio]&=\Erm[\Erm[\bt^i|\zio]|\xio]\\
    &=\Erm[\bM^{-1}_{\indexO_i}\Wiobs^\intercal\zio|\xio]\\
    &=\bM^{-1}_{\indexO_i}\Wiobs^\intercal\Erm[\zio|\xio].
\end{align*}
Then
\begin{align*}
    \Erm[\bt^i(\zio)^\intercal|\xio]&= \Erm[\Erm[\bt^i(\zio)^\intercal|\zio]|\xio]\\
    &=\Erm[\Erm[\bt^i|\zio](\zio)^\intercal|\xio]\\
    &=\bM^{-1}_{\indexO_i}\Wiobs^\intercal\Erm[\zio(\zio)^\intercal|\xio]\\
    &=\bM^{-1}_{\indexO_i}\Wiobs^\intercal
    \left(\Covrm[\zio|\xio]+\Erm[\zio|\xio]\Erm[(\zio)^\intercal|\xio]\right)\\
    &=\bM^{-1}_{\indexO_i}\Wiobs^\intercal\Covrm[\zio|\xio]+
    \Erm[\bt^i|\xio]\Erm[(\zio)^\intercal|\xio].
\end{align*}
and 
\begin{align*}
    &\Erm[\bt^i(\bt^i)^\intercal|\xio]= \Erm[\Erm[\bt^i(\bt^i)^\intercal|\zio]|\xio]
    \\
    =&\Erm[\bM^{-1}_{\indexO_i}\Wiobs^\intercal\zio (\zio)^\intercal \Wiobs \bM^{-1}_{\indexO_i} |\xio]+\Erm[\Covrm[\bt^i|\zio]|\xio]
    \\
    =&\bM^{-1}_{\indexO_i}\Wiobs^\intercal \Erm[\zio (\zio)^\intercal|\xio] \Wiobs \bM^{-1}_{\indexO_i}
    + \Erm[\sigma^2\bM^{-1}_{\indexO_i}|\xio]
    \\
    =&\bM^{-1}_{\indexO_i}\Wiobs^\intercal \left(\Covrm[\zio|\xio]+\Erm[\zio|\xio]\Erm[(\zio)^\intercal|\xio]\right) \Wiobs \bM^{-1}_{\indexO_i}
    + \sigma^2\bM^{-1}_{\indexO_i}
    \\
    =&\bM^{-1}_{\indexO_i}\Wiobs^\intercal \Covrm[\zio|\xio] \Wiobs \bM^{-1}_{\indexO_i} + 
    \Erm[\bt^i|\xio] \Erm[(\bt^i)^\intercal|\xio] +
    \sigma^2\bM^{-1}_{\indexO_i}.
\end{align*}

\subsection{M-step details}
Take the derivative of the Q-function in \cref{Eq;Q-func} with respect to row $\bw_j^\intercal$ and $\sigma^2$:
\begin{align*}
    \frac{\partial Q}{\partial \bw_j^\intercal}&=
    \frac{-1}{|\Omega_j|\sigma^2}\sum_{i\in \Omega_j}(-\be_j^\intercal\Ebb[\zio\bt_i^\intercal] + \bw_j^\intercal \Ebb[\bt_i\bt_i^\intercal]),
    \nonumber\\
    \frac{\partial Q}{\partial \sigma^2}&=\frac{1}{2\sigma^4}\sum_{i=1}^n
   \left(\Ebb[(\zio)^\intercal \zio] - 2\tr(\Wiobs\Ebb[\bt_i(\zio)^\intercal]) +  \tr(\Wiobs^\intercal \Wiobs\Ebb[\bt_i\bt_i^\intercal])\right)
   -\frac{\sum_{i=1}^n|\indiObs|}{2\sigma^2}.\nonumber
\end{align*}
Set both to zero to obtain the update for M-step:
\begin{align*}
   \hat \bw_j^\intercal&=\left(\be_j^\intercal\sum_{i\in \Omega_j}\Ebb[\zio\bt_i^\intercal] \right)
   \left(\sum_{i\in \Omega_j}\Ebb[\bt_i\bt_i^\intercal]\right)^{-1},\\
    \hat{\sigma}^2
    &=\frac{1}{\sum_{i=1}^n|\indiObs|}\sum_{i=1}^n
    \left(\Ebb[(\zio)^\intercal \zio] - 2\tr(\hat \bW_{\indexO_i}\Ebb[\bt_i(\zio)^\intercal]) +  \tr(\hat \bW_{\indexO_i}^\intercal \hat \bW_{\indexO_i}\Ebb[\bt_i\bt_i^\intercal])\right).
\end{align*}

\subsection{Approximation of the truncated normal moments}
The region $g_j^{-1}(\xij)$ is an interval:  $g_j^{-1}(\xij)=(a_{ij},b_{ij})$. We may consider three cases: (1) $a_{ij},b_{ij} \in \mathbb{R}$; (2) $a_{ij}\in \mathbb{R}, b_{ij}=\infty$; (3) $a_{ij}=-\infty, b_{ij}\in \mathbb{R}$. The computation for all cases are similar. We take the first case as an example. First we introduce a lemma for a univariate truncated normal.

\begin{lemma}
	Consider a univariate random variable $z\sim \mathcal{N}(\mu,\sigma^2)$. For constants $a<b$, let $\alpha=(a-\mu)/\sigma$ and $\beta=(b-\mu)/\sigma$. Then the mean and variance of $z$ truncated to the interval $(a,b)$ are:
	\begin{equation*}
		\Erm(z|a<z\leq b) = \mu +\frac{\phi(\alpha)-\phi(\beta)}{\Phi(\beta)-\Phi(\alpha)} \cdot \sigma,
		\label{Eq:mean_tmean}
	\end{equation*} 
		\begin{equation*}
	\Varrm (z|a<z\leq b) =\left( 1+ \frac{\alpha\phi(\alpha)-\beta\phi(\beta)}{\Phi(\beta)-\Phi(\alpha)}- \left(\frac{\phi(\alpha)-\phi(\beta)}{\Phi(\beta)-\Phi(\alpha)} \right)^2  \right)\sigma^2 := h(\alpha, \beta,\sigma^2).
		\label{Eq:mean_tvar}
	\end{equation*} 
	\label{lemma:truncated}
\end{lemma}

Notice conditional on known $\bz^i_{\indexO_i/\{j\}}$, $\zij$ is normal with mean $\mu_{ij}$ and variance $\sigma_{ij}^2$ as
\begin{align}
        \mu_{ij}&=\bw_j^\intercal (\sigma^2\Irm_k+\bW_{\indexO_i/\{j\}}^\intercal \bW_{\indexO_i/\{j\}})^{-1}\bW_{\indexO_i/\{j\}}\bz^i_{\indexO_i/\{j\}},
        \label{Eq:muij}\\
            \sigma_{ij}^2 &= \sigma^2 + \sigma^2 (\sigma^2\Irm_k+\bW_{\indexO_i/\{j\}}^\intercal \bW_{\indexO_i/\{j\}})^{-1}\bw_j^\intercal \bw_j.
            \label{Eq:sigmaij}
\end{align}
Now define $\alpha_{ij}=\frac{a_{ij}-\mu_{ij}}{\sigma_{ij}}$ and $\beta_{ij}=\frac{b_{ij}-\mu_{ij}}{\sigma_{ij}}$ as in \cref{lemma:truncated}.

We first discuss how to estiamte $\Erm[\zio|\xio]$.
Again using the law of total expectation for each $j\in\indexO_i$ by treating $ \bz^i_{\indexO_i/\{j\}}$ as known:
\begin{align}
    &\Erm[\zij|\xio] = \Erm[ \Erm[\zij|\xij,\bz^i_{\indexO_i/\{j\}}]|\xio]
    \nonumber 
    =\Erm\left[ \mu_{ij}
    -\frac{\phi(\alpha_{ij})-\phi(\beta_{ij})}{\Phi(\beta_{ij})-\Phi(\alpha_{ij})} \sigma_{ij}|\xio\right]
    \nonumber \\
    =& \bw_j^\intercal (\sigma^2\Irm+\bW_{\indexO_i/\{j\}}^\intercal \bW_{\indexO_i/\{j\}})^{-1}\bW_{\indexO_i/\{j\}}
    \Erm[\bz^i_{\indexO_i/\{j\}}|\xio]
    -\Erm\left[\frac{\phi(\alpha_{ij})-\phi(\beta_{ij})}{\Phi(\beta_{ij})-\Phi(\alpha_{ij})}\large|\xio\right] \sigma_{ij}.
    \label{Eq:system_1}
\end{align}
Notice $\Erm\left[\frac{\phi(\alpha_{ij})-\phi(\beta_{ij})}{\Phi(\beta_{ij})-\Phi(\alpha_{ij})}\large|\xio\right]$ is the expectation of a nonlinear function of $\bz^i_{\indexO_i/\{j\}}$ with respect to the conditional distribution $\bz^i_{\indexO_i/\{j\}}|\xio$.
Such expectation is intractable, thus we resort to an linear approximation:
\begin{align}
        \Erm\left[\frac{\phi(\alpha_{ij})-\phi(\beta_{ij})}{\Phi(\beta_{ij})-\Phi(\alpha_{ij})}\large|\xio\right] \approx \frac{\phi(\Erm[\alpha_{ij}|\xio])-\phi(\Erm[\beta_{ij}|\xio])}{\Phi(\Erm[\beta_{ij}|\xio])-\Phi(\Erm[\alpha_{ij}|\xio])}.
        \label{Eq:approx_1}
\end{align}
where $\Erm[\alpha_{ij}|\xio]$ and $\Erm[\beta_{ij}|\xio]$ are linear functions of $\Erm[\bz^i_{\indexO_i/\{j\}}|\xio]$.

Combining \cref{Eq:system_1} and \cref{Eq:approx_1},
we approximately express the $j$-th element of $\Erm[\zio|\xio]$ as a nonlinear function (including a linear part) of all other elements of $\Erm[\zio|\xio]$.
Such relationship holds for all $j\in \indexO_i$,
thus we have a system with $|\indexO_i|$ equations satisfied by the vector $\Erm[\zio|\xio]$.

We choose to iteratively solve this system.
Concretely, 
to estimate $\Erm[\zio|\xio,\bW^{(t+1)},(\sigma^2)^{(t+1)}]$  at the $t+1$-th EM iteration,
we conduct one Jacobi iteration with $\Erm[\zio|\xio,\bW^{(t)},(\sigma^2)^{(t)}]$ as initial value.
Surprisingly, one Jacobi iteration works well and more iterations do not bring significant improvement.

The values of $\mu_{ij}$ and $\sigma_{ij}^2$ in \cref{Eq:muij} and \cref{Eq:sigmaij} for all $j\in\indexO_i$ can be obtained through computing the diagonals of $(\sigma^2\Irm_{|\indiObs|}+\ \bW_{\indexO_i}\bW_{\indexO_i}^\intercal)^{-1}$ and $(\sigma^2\Irm_{|\indiObs|}+\ \bW_{\indexO_i}\bW_{\indexO_i}^\intercal)^{-1}\Erm[\zio|\xio,\bW^{(t)},(\sigma^2)^{(t)}]$,
which makes the computation no more than $O(k^2|\indexO_i|)$ for each data point at each EM iteration.

As for diagonals of $\Covrm[\zio|\xio]$, denoted as $\Varrm[\zio|\xio]\in \Rbb^{|\indexO_i|}$, 
using the law of total variance,
\begin{equation}
    \Varrm[\zij|\xio]=\Erm[\Varrm[\zij|\bz^i_{\indexO_i/\{j\}},\xij]|\xio]]+\Varrm[\Erm[\zij|\bz^i_{\indexO_i/\{j\}},\xij]|\xio].
    \label{Eq:approx_2}
\end{equation}
In the right hand side of \cref{Eq:approx_2},
we similarly approximate the first term, an intractable non-linear integral, as a linear term:
\begin{equation}
    \Erm[\Varrm[\zij|\bz^i_{\indexO_i/\{j\}},\xij]|\xio] \approx h(\Erm[\alpha_{ij}|\xio], \Erm[\beta_{ij}|\xio], \sigma^2_{ij}).
    \label{Eq:approx_3}
\end{equation}
The second term in the right hand side of \cref{Eq:approx_2} is also an intractable nonlinear integral.
We approximate it as $0$ and find it works well then linearly approximating it in practice.

Combining \cref{Eq:approx_2} and \cref{Eq:approx_3} for all $j\in\indexO_i$, 
we express $\Varrm[\zio|\xio]$ by nonlinear functions of $\Erm[\zio|\xio]$.
Thus at the $t+1$-th EM iteration,
we first estimate $\Erm[\zio|\xio,\bW^{(t+1)},(\sigma^2)^{(t+1)}]$,
then estimate $\Varrm[\zio|\xio]$ based on that.

\subsection{Stopping criteria}
We use the relative change of the parameter $\bW$ as the stopping criterion.
Concretely, with $\bW_1$ from last iteration and $\bW_2$ from current iteration, 
the algorithm stops if $\frac{||\bW_1-\bW_2||_F^2}{||\bW_1||_F^2}$ is smaller than the tolerance level.

\section{Additional experiments}
\subsection{LRGC imputation under correct model}
For LRGC imputation,
we show the random variation of the error (due to error in the estimate of $\zmis$) dominates the estimation error (due to errors in the estimates of the parameters $\bW$ and $\sigma$).
To do so, we compare the imputation error of LRGC imputation with estimated model parameters (\texttt{LRGC}) and true model parameters (\texttt{LRGC-Oracle}).
For ordinal data,
imputation requires approximating truncated normal moments, which may blur the improvement of using true model parameters.
Thus we conduct the comparison on the same continuous synthetic dataset described in Section 4.
The results are reported in \cref{table:LRGC_oracle}.

\begin{table}
	\centering
	\caption{Imputation error (NRMSE) on synthetic continuous data over $20$ repetitions.}
	\label{table:LRGC_oracle}
	\begin{tabular}{lcc}
		\toprule
		Setting & \texttt{LRGC}  & \texttt{LRGC-Oracle}   \\
		\midrule
		Low Rank & $0.347(.004))$  & $.330(.004)$  \\
		High Rank & $0.517(.011)$  & $.433(.007)$  \\
		\bottomrule
	\end{tabular}
\end{table}

Compared to \texttt{LRGC},
\texttt{LRGC-Oracle} only improves slightly ($1\%$) over low rank data.
Thus the model estimation error is dominated by the random variation of the imputation error.
For high rank data,
the improvement ($8\%$) is still small compared to the gap between LRGC imputation and LRMC algorithms ($\geq 18\%$).
Also notice the marginal transformation $g_j(z)=z^3$ for high rank data is not Lipschitz,
so the theory presented in this paper does not bound the LRGC imputation error.

The result here indicates there is still room to improve LRGC imputation when 
the marginals are not Lipschitz.
We leave that important work for the future.

\subsection{Imputation error versus reliability shape with varying number of ordinal levels}
We show in this section the imputation error versus reliability curve shape on ordinal data with many ordinal levels will match that on continuous data.
The results here indicate the prediction power of \texttt{LRGC} reliability depends on the imputation task. 
The prediction power is larger for easier imputation task.
In the synthetic experiments, 
imputing continuous data is harder than imputing ordinal data,
and imputing 1-5 ordinal data is harder than imputing binary data.

\begin{figure}
 	\centering
 	\includegraphics[width=0.8\linewidth]{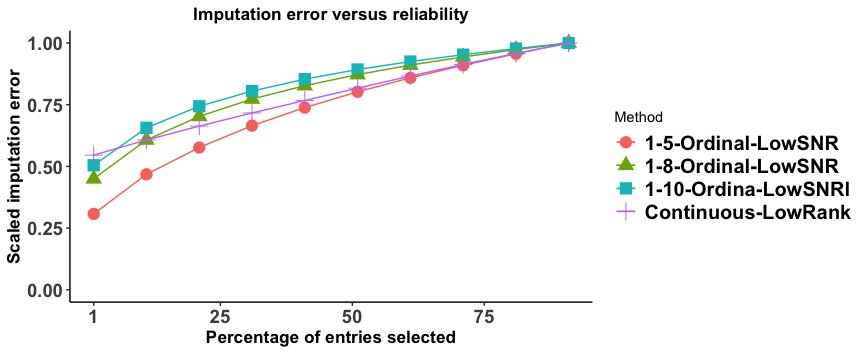}
 	\caption{Imputation error on the subset of $m\%$ most reliable entries, reported over $5$ repetitions.}
 	\label{fig:uncertainty_varying}
 \end{figure}

We follow the synthetic experiments setting used in Section 4, but vary the number of ordinal levels to $\{5,8,10\}$.
We adopt high SNR setting for ordinal data and low rank setting for continuous data.
To make the imputation error comparable between continuous data and ordinal data,
we measure the ratio of the imputation error over the $m\%$ entries to the imputation error over all missing entries.
Shown in \cref{fig:uncertainty_varying},
the curve shape for low rank continuous data is similar to that for ordinal data with 5--8 levels.
Also notice, NRMSE for continuous data involves the observed data values while MAE does not,
which may cause small difference in the curve shape.

\section{Experimental detail}
\subsection{Synthetic data}

To select the best value of the key tuning parameter for each method, 
we first run some initial experiments to determine a proper range such that the best value lies strictly inside that range.

For \texttt{LRGC}, the only tuning parameter is rank.
We find that a range of $6-14$  for continuous data (both low and high rank),
and $3-11$ for ordinal data with 5 levels and binary data (high SNR and low SNR),
suffices to ensure the best value is strictly inside the range. 
Notice this range is still quite small, so it is rather easy to search over.

For \texttt{softImpute}, the only tuning parameter is the penalization parameter.
As suggested by the vignette of the R package \cite{hastie2015softimpute},
we first center the rows and columns of the observations using the function $\texttt{biScale()}$ and 
then compute $\lambda_0$ as an upper bound on the penalization parameter using the function $\texttt{lambda0()}$.
The penalization parameter range is set as the exponentially decaying path between $\lambda_0$ and $\lambda_0/100$ with nine points for all cases:
\begin{verbatim}
  exp(seq(from=log(lam0),to=log(lam0/100),length=9)).
 \end{verbatim}
 We found increasing the path length from $9$ to $20$ only slightly improves the performance (up to $.01$ across all cases) on best performance on test set.

 For \texttt{MMMF}, there are two tuning parameters: the rank and the penalization parameter. 
 We set the rank to be allowed maximum rank $199$.
 We set the range of penalization parameter as we do for \texttt{softImpute},
 with left and right endpoints that depend on the data.
 For \texttt{MMMF}-$\ell_2$ on continuous data and \texttt{MMMF-BvS} on ordinal data, we use $\lambda_0/4$ as start point and $\lambda_0/100$ as end point.
 For all other \texttt{MMMF} methods, we use $\lambda_0$ as start point and $\lambda_0/100$ as end point.
 
 For \texttt{MMC}, following the authors' suggestions regarding the code,
 we use the following settings: 
 (1) the number of gradient steps used to update the $Z$ matrix is $1$;
 (2) the tolerance parameter is set as $0.01$;
 In addition,
 we set the initial rank as $50$, 
 the increased rank at each step as $5$, 
 the maximum rank as $199$,
 the maximum number of iterations as $80$ and
 the Lipshitz constant as $10$.
 Finally, the key tuning parameter we search over is the constant step size as suggested by the authors of \cite{ganti2015matrix}.
 The range is set as $\{3,5,7,\ldots,17,19\}$.

\begin{figure}
 	\centering
 	\includegraphics[width=\linewidth]{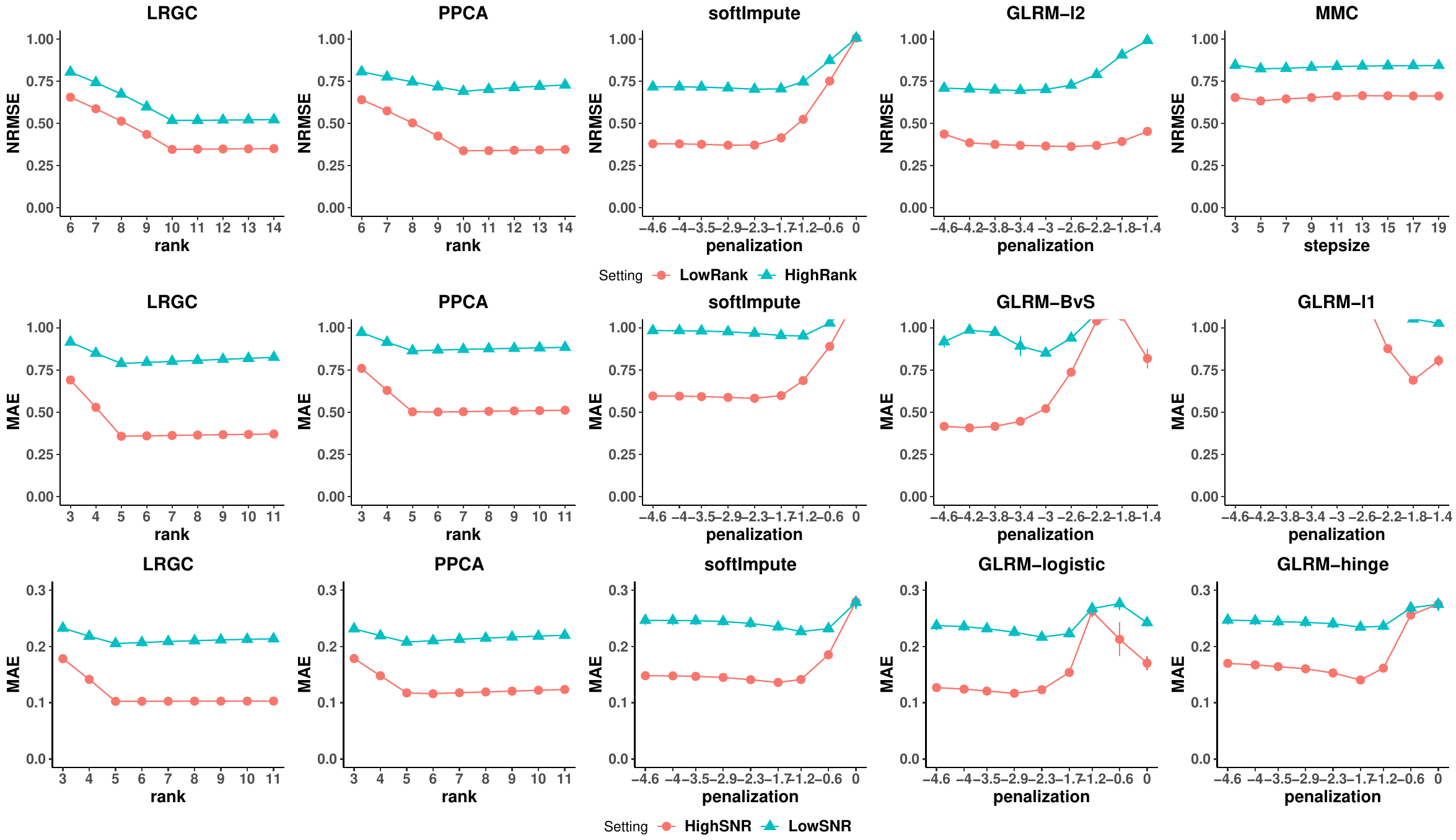}
 	\caption{Imputation error over a key tuning parameter reported over $20$ repetitions. The error bars ara invisible. 
 	The penalization parameter $\lambda$ is plotted over the log-ratios $\log(\alpha)$ which satisfies $\lambda = \alpha \lambda_0$.}
 	\label{fig:error_sim}
 \end{figure}

 The complete results are plotted in \cref{fig:error_sim}. 
 Clearly, \texttt{LRGC} does not overfit even for high ranks, across all settings.
 We also provide the runtime for each method at the best tuning parameter in \cref{table:simulation_runtime}.
 Notice our current implementation is written entirely in R, 
 and thus further acceleration is possible.

 \begin{table}
	\centering
	\caption{Run time (in seconds) for synthetic data at the best tuning parameter; mean (variance) reported over 20 repetitions.}
	\label{table:simulation_runtime}
	\begin{tabular}{lccccc}
		\toprule
		Continuous  & LRGC & PPCA   & softImpute  & MMMF-$\ell_2$ & MMC\\
		\midrule
		Low Rank  & $5.7(0.2)$& $2.9(0.4)$  & $0.7(0.0)$   &$3.3(1.0)$ & $457.9(10.4)$\\
	High Rank     & $6.5(0.3)$ & $0.3(0.1)$ & $1.1(0.2)$  & $7.6(2.0)$ & $554.4(32.0)$\\
	\midrule
	1-5 ordinal  & LRGC & PPCA & softImpute & MMMF-BvS & MMMF-$\ell_1$\\
	\midrule
		High SNR   & $27.2(0.7)$& $1.0(0.2)$ & $1.2(0.1)$ & $19.2(1.5)$ & $17.4(1.2)$\\
		Low SNR  & $19.8(0.8)$& $0.3(0.1)$  & $1.3(0.0)$   &$17.4(1.5)$ &  $17.0(1.4)$\\
		\midrule
		Binary	& LRGC & PPCA & MMMF-hinge & MMMF-logistic & MMMF-$\ell_1$\\
			\midrule
High SNR    & $66.7(3.0)$ & $0.9(0.5)$ &$3.8(0.3)$  & $4.4(0.6)$ & $2.1(0.3)$\\
Low SNR   & $52.0(4.4)$ &$0.3(0.1)$ &$3.4(0.4)$  & $3.3(0.4)$ & $1.9(0.2)$\\
		\bottomrule
	\end{tabular}
\end{table}

\subsection{MovieLens 1M}
The dataset can be found at \url{https://grouplens.org/datasets/movielens/1m/}.
Similar to the synthetic experiments, we choose the tuning parameter for each method on a proper range determined through some initial experiments.
For \texttt{LRGC}, we choose the rank from $\{8,10,12,14\}$ to be $10$.
With $\lambda_0$ calculated as for the synthetic data,
for \texttt{softImpute}, we select the penalization parameter from  $\{\frac{\lambda_0}{2},\frac{\lambda_0}{4},\frac{\lambda_0}{6},\frac{\lambda_0}{8}\}$ to be $\frac{\lambda_0}{4}$;
for \texttt{MMMF-BvS}, we set the rank as $200$ and select the penalization parameter from $\{\frac{\lambda_0}{10},\frac{\lambda_0}{12},\frac{\lambda_0}{14},\frac{\lambda_0}{16},\frac{\lambda_0}{18}\}$ to be $\frac{\lambda_0}{14}$.

We report detailed results in \cref{table:movielens}.
We see that all the models perform quite similarly on this large dataset.
In other words, the gain from carefully modeling the marginal distributions
(using a LRGC) is insignificant. 
This phenomenon is perhaps unsurprising given that sufficiently large
data matrices from a large class of generative models are approximately low rank \cite{udell2019why}.

\begin{table}
	\centering
	\caption{Imputation error  for MovieLens 1M over $5$ repetition. Run time is measures in minutes.}
	\label{table:movielens}
	\begin{tabular}{lcccc}
		\toprule
		Algorithm & MAE   & RMSE  & Run time \\
		\midrule
		LRGC & $\mathbf{0.619(.002)}$  & $0.910(.003)$   &$38(1)$ \\
		softImpute  & $0.629(.003)$  & $\mathbf{0.905(.003)}$   &$93(2)$ \\
MMMF-BvS & $0.633(.002)$ & $0.921(.002)$  & $25(1)$ \\
		\bottomrule
	\end{tabular}
\end{table}


\end{appendix}

\end{document}